\documentclass[pdflatex,sn-mathphys]{sn-jnl}
\jyear{2021}%
\theoremstyle{thmstyleone}

\theoremstyle{thmstyletwo}%

\theoremstyle{thmstylethree}%

\usepackage[labelsep=period]{caption}
\renewcommand{\figurename}{Fig.}
\usepackage{multirow}
\usepackage{multicol}
\usepackage{graphicx}
\usepackage{array}
\usepackage{arydshln}
\usepackage{color}
\usepackage{adjustbox}
\usepackage{amssymb}
\usepackage{tabularx}
\usepackage{float}
\usepackage{lineno}
\usepackage{blindtext}
\usepackage{stfloats}
\usepackage{makecell}
\usepackage{graphicx} 
\usepackage{bm}
\usepackage{mathrsfs}
\usepackage{amsmath}
\usepackage{threeparttable}
\usepackage{xcolor}
\usepackage{url}

\let\proglang=\textsf

\algnewcommand{\Input}[1]{%
  \State \textbf{Input:}
  \Statex \hspace*{\algorithmicindent}\parbox[t]{.8\linewidth}{\raggedright #1}
}
\algnewcommand{\Initialize}[1]{%
  \State \textbf{Initialize:}
  \Statex \hspace*{\algorithmicindent}\parbox[t]{.8\linewidth}{\raggedright #1}
}
\algnewcommand{\Output}[1]{%
  \State \textbf{Output:}
  \Statex \hspace*{\algorithmicindent}\parbox[t]{.8\linewidth}{\raggedright #1}
}

\begin{document}

\title[A hybrid ensemble method with negative correlation learning for regression]{A hybrid ensemble method with negative correlation learning for regression}

\author[1]{\fnm{Yun} \sur{Bai}}\email{baiyun12138@buaa.edu.cn}

\author[2]{\fnm{Ganglin} \sur{Tian}}\email{ganglin.tian@imt-atlantique.net}

\author*[1]{\fnm{Yanfei} \sur{Kang}}\email{yanfeikang@buaa.edu.cn}

\author[1]{\fnm{Suling} \sur{Jia}}\email{jiasuling@buaa.edu.cn}

\affil*[1]{\orgdiv{School of Economics and Management}, \orgname{Beihang University}, \orgaddress{\city{Beijing}, \postcode{100191}, \country{China}}}

\affil[2]{\orgdiv{Faculty of Microwave, Observation, and Perspection of Environment}, \orgname{IMT Atlantique}, \orgaddress{\city{Plouzané}, \postcode{29280}, \country{France}}}

\abstract{Hybrid ensemble, an essential branch of ensembles, has flourished in the regression field, with studies confirming diversity's importance.
However, previous ensembles consider diversity in the sub-model training stage, with limited improvement compared to single models.
In contrast, this study automatically selects and weights sub-models from a heterogeneous model pool.
It solves an optimization problem using an interior-point filtering linear-search algorithm.
The objective function innovatively incorporates negative correlation learning as a penalty term, with which a diverse model subset can be selected.  
The best sub-models from each model class are selected to build the NCL ensemble, which performance is better than the simple average and other state-of-the-art weighting methods.
It is also possible to improve the NCL ensemble with a regularization term in the objective function.
In practice, it is difficult to conclude the optimal sub-model for a dataset prior due to the model uncertainty. 
Regardless, our method would achieve comparable accuracy as the potential optimal sub-models.
In conclusion, the value of this study lies in its ease of use and effectiveness, allowing the hybrid ensemble to embrace diversity and accuracy.}

\maketitle

\section{Introduction}
Ensemble learning has been proven to be theoretically and empirically superior to single models by state-of-the-art literature as a method of combining pre-trained models in a certain way to obtain final predictions \cite{brown2005diversity,chandra2006evolving,mendes2012ensemble}.
Typically, ensemble models sample the input space of data and features, such as cross-validation or down-sampling of data \citep{leblanc1996combining}. 
Meanwhile, features can be selected by calculating feature importance \citep{mendes2012ensemble}. 
Then, ensembles are followed by combining multiple but homogeneous weak learners to form a strong learner to achieve higher accuracy.
The famous examples of ensemble models are bagging \citep{breiman1996bagging}, boosting \citep{freund1996experiments}, and stacking \citep{wolpert1992stacked}.
In recent years, solutions based on ensemble models often achieve good results in \href{https://www.kaggle.com/}{\emph{Kaggle competitions}} \cite{taieb2014gradient,hoch2015ensemble,bojer2020kaggle}.

In some pioneering studies, researchers attempted to train completely heterogeneous models for the same input space and then averaged or weighted the predictions of these models.
This approach considered that heterogeneous models were more likely to increase diversity during training and produce more robust results compared to homogeneous models \cite{zhao2010incremental,mendes2012ensemble}.
Training with heterogeneous models also refers to the hybrid ensemble.
For load prediction, Salgado chose several support vector machines and neural networks, ranked and filtered the candidates, and finally weighted the predictions of the selected models. 
Their hybrid ensemble model improved performance by 25\% over the best single predictor \cite{salgado2006hybrid}.
Ala'raj took five classifiers and combined their predictions. 
The experimental results demonstrated the ability of the proposed method to improve the accuracy of credit scoring prediction \cite{ala2016new}.
Qi constructed a hybrid ensemble model for predicting slope stability in geology, which included six sub-models, such as support vector machines and artificial neural networks.
A genetic algorithm was introduced to calculate the classification weights for each model. 
This hybrid ensemble outperformed any single model, even though the single model already had its optimal parameters \cite{qi2018hybrid}.
Some researchers constructed ensembles containing both homogeneous and heterogeneous models.
For example, Merz chose six multivariate adaptive regression splines and six back-propagation networks to build a model pool, ranked the sub-models by principal components with the variance from the learning process to highlight the contributions of different sub-models \cite{merz1999principal}.

Scholars have identified model diversity as a critical factor to hybrid ensemble success \cite{brown2004diversity,webb2004multistrategy,chandra2006evolving}.
In recent years, researchers have put effort into ensemble diversity and generalization.
The authors developed a pruning method for classification ensembles utilizing the tradeoff between accuracy and diversity \cite{bian2021does}.
Several methods to increase the diversity of sub-models within an ensemble are also proposed. 
For earlier schemes, practitioners trained models with cross-validation or chose different parameter combinations for homogeneous models, followed by majority voting or weighted averaging of the model predictions.
Cross-validation yet provided limited improvement for model accuracy, and Stone proved as early as 1974 that estimators generated by cross-validation behaved similarly \cite{stone1974cross}.
Hansen and Salomon proposed using neural networks to construct ensembles in the 1990s. 
They used neural networks to fit different parts of the training data, which were then majority voted as the result of ensemble \cite{hansen1990neural}.
Both Ting and Cano obtained a diversity of sub-models by using different subsets of features \cite{ting2011feature,cano2020kappa}. 
Ting emphasized that unstable learners could generate sufficient diversity of global models since they were more sensitive to data changes \cite{ting2011feature}.
Cano suggested dynamically monitoring the model pool to eliminate the oldest and weakest sub-models in time for the streaming data scenario \cite{cano2020kappa}.
Sirovetnukul pointed out that a hybrid ensemble could learn negative knowledge from less well-performed models that were easily ignored and removed in previous studies. 
Such knowledge could help the models converge to better solutions while producing diverse results \cite{sirovetnukul2011effectiveness}.
Brown considered the negative knowledge across sub-models and provided quantitative methods for the diversity of hybrid ensembles \cite{brown2005diversity}.

Some empirical evidence demonstrated the ability of Negative Correlation learning (NCL) to increase model diversity and improve ensemble models \cite{liu1999ensemble,liu2000evolutionary,chandra2006evolving,sirovetnukul2011effectiveness,alhamdoosh2014fast,peng2020negative}.
NCL introduces a correlation penalty term in the objective function of each sub-model to measure the deviation from the current ensemble.
All sub-models can be trained simultaneously and interactively on the same training set, and the final experimental results will achieve a bias-variance-covariance balance, as theoretically deduced.
Current applications of NCL are focused primarily on the training process of ensemble neural networks to diversify each sub-model \cite{liu1999ensemble,liu2000evolutionary,tang2009selective,alhamdoosh2014fast,hadavandi2015novel,peng2020negative}.
Although the ensemble neural network trains the sub-models with diversity under NCL, they are still structurally homogeneous models, differing only in specific parameters.
To our knowledge, only some studies apply NCL to hybrid ensembles.
Next, we will discuss the feasibility of using NCL to improve hybrid ensembles.

Generally, ensembles contain two stages: sub-model training and combination \cite{merz1999using}.
Previous ensembles used NCL as a penalty term to train diverse sub-models in the first stage, followed by some basic methods, such as majority voting or simple averaging, to combine the predictions, ignoring the role of diversity in the second stage.
In contrast, the hybrid ensemble trains multiple heterogeneous models based on the consensus that heterogeneous models will produce diverse predictions in the first stage \cite{zhao2010incremental,mendes2012ensemble}.
In the second stage, if we apply NCL to the objective function to optimize the weights of each sub-model, it is possible to select a diverse set of sub-models to obtain the final results.
We present the methods for obtaining diversity at different stages of the ensemble models in Table~\ref{Methods for obtaining diversity at different stages of the ensemble model.} for comparison.

\begin{table}[htbp]
\caption{Methods for obtaining diversity at different stages of the ensemble model.}
  \label{Methods for obtaining diversity at different stages of the ensemble model.}
\begin{tabular}{lll}\\
\toprule
Ensemble models                                                     & Stage 1: sub-models training
  & Stage 2: sub-models combination
  \\
  \midrule
\begin{tabular}[c]{@{}l@{}}Ensemble neural \\ networks\end{tabular} & \begin{tabular}[c]{@{}l@{}}Homogeneous sub-models are \\ trained simultaneously and \\ interactively to increase the \\ diversity of sub-models \\ during the training process.\end{tabular} & \begin{tabular}[c]{@{}l@{}}Majority voting or simple \\ averaging is used to combine \\ the predictions, not considering \\ the diversity within the ensemble.\end{tabular} \\ \midrule
Hybrid ensembles                                                    & \begin{tabular}[c]{@{}l@{}}Heterogeneous models are \\ trained separately to ensure \\ diversity.\end{tabular}      
& \begin{tabular}[c]{@{}l@{}}A diverse subset of predictions is \\ selected and weighted by NCL.\end{tabular}                                                     \\ \midrule
\end{tabular}
\end{table}

To improve hybrid ensembles with NCL, we design a generic scheme in this study for regression problems.
Eleven well-established regression prediction methods, including ensemble and generalized linear regression models, are fed to the model pool.
Each sub-model is trained and generates a set of predictions.
Cross-validation and grid search are applied to the training process to obtain the predictor with the optimal parameters.
Subsequently, we view the process of the second stage of hybrid ensembles, sub-model combination, as an optimization problem.
This problem can be solved using the interior-point filter line-search algorithm \cite{wachter2006implementation}, which is a solver in the \proglang{Gekko} optimizer developed by Beal \cite{beal2018gekko}.
We add NCL as a penalty term to the objective function of the optimization problem.
We designed several experiments to evaluate the proposed method from multiple dimensions.
The hybrid ensemble for regression based on NCL achieves excellent results, demonstrating its great potential.

The main contributions of this study are three-fold:
\begin{itemize}

\item[(1)] Initially, this study attempts to migrate the application scenario of NCL from the traditional sub-model training stage to the sub-model combination stage, with good results in a hybrid ensemble consisting of heterogeneous sub-models.

\item[(2)] The model selection and combination process is treated as an optimization problem. This problem leads to a diverse set of sub-models in the model pool, given by a weight vector.

\item[(3)] Ultimately, the approach in this study again verifies that diversity is the key to the success of ensemble models, and it is an innovation to ensure model diversity in both stages of the hybrid ensemble.

\end{itemize}

The rest of this paper is organized as follows. Section~\ref{Related works} introduces the theories and methods involved in the proposed framework.
Section~\ref{Hybrid ensemble for regression with negative correlation learning} presents a hybrid ensemble based on NCL, accounting for model diversity. In Section~\ref{Experiments}, we systematically investigate the application of the proposed method on twenty publicly available datasets and analyze the contribution of NCL to performance improvement. Section~\ref{Discussion} reviews the background of our proposed method, illustrates the method's ability to remedy some of the shortcomings of current hybrid ensemble studies, and synthesizes the experimental performance and scope for improvement of our method. Finally,  Section~\ref{Conclusion} concludes the paper.

\newpage
\section{Related works}
\label{Related works}

This section first introduces ambiguity and bias-variance-covariance decompositions, which are the theoretical basis for Negative Correlation Learning (NCL) to increase the diversity of hybrid ensembles \cite{brown2005diversity}. 
The general form of the NCL is presented in the second part. 
The third part shows the computational principles and applications of the interior-point filter line-search algorithm.

\subsection{Two types of decomposition}
In the context of multiple regression, there is a dataset containing \emph{n} samples with $\{(x_1,y_1),...,(x_n,y_n)\}$. The objective of the problem is to find a function \emph{f} that maps $\mathbb{R}^{n}$ to $\mathbb{R}^{1}$ to gain predictive capability for future data.
In machine learning, \emph{f} is a model or an estimator.
\begin{equation}
    \emph{f}(x_i) = y_i,\quad \quad \emph{f}:  \mathbb{R}^{n} \rightarrow \mathbb{R}^{1}, x_i \in \mathbb{R}^{n},  y_i \in \mathbb{R}^{1}.
\end{equation}

\subsubsection{Ambiguity decomposition}
In a general scenario, \emph{m} sub-models can form a hybrid ensemble $f_{h}$ with a weighted average. 
$f_{h}$ is a convex combination of all components:
\begin{equation}
    f_{h} = \sum_{j=1}^{m} \omega_j f_j,
\end{equation}
where $\sum_{j=1}^{m} \omega_j = 1$, and $f_j$ is the predictions of $j_{th}$ sub-model.
According to Brown, the Mean Square Error (MSE) $\zeta_h$ of $f_h$ can be expressed as the difference between the following two terms \cite{brown2005diversity}:
\begin{equation}
\label{MSE_HE}
    \zeta_h = \sum_{j=1}^{m}\omega_j\zeta_j-\frac{1}{n}\sum_{j=1}^{m}\sum_{i=1}^{n}\omega_j(f_h(x_i)-f_j(x_i))^2,
\end{equation}
where $\zeta_j = \frac{1}{n}\sum_{i=1}^{n}\left( f_j(x_i) - y_i \right)^2$.
The first term of Equation~(\ref{MSE_HE}) is the weighted average of the MSE of each sub-model; the second is the ambiguity term.
Equation~(\ref{MSE_HE}) indicates that $\zeta_h$ is less than the weighted average $\zeta_j$ of all sub-models, given that the sub-models are not identical and the second ambiguity term is positive.
This fact reveals that the more significant the difference between each sub-model and the current hybrid ensemble, the larger the ambiguity term and the smaller the MSE of the hybrid ensemble.
Notably, without an established criterion to judge the best model in advance, it is efficient to use the hybrid ensemble directly, even if some member has the lowest error.

\subsubsection{Bias-variance-covariance decomposition}
The MSE of the sub-models and the hybrid ensemble are employed in the ambiguity decomposition to measure diversity; the higher the second term in Equation~(\ref{MSE_HE}), the more diverse the ensemble.
However, as the sub-models increase in volume, they are more likely to deviate from the actual value, although they would get more diverse.
This situation leads to an increase in the first term of $\zeta_h$ when it is not so beneficial to consider increasing the diversity of the hybrid ensemble.
Thus, balancing the diversity and accuracy of the sub-models and ensemble is of interest. 
The bias-variance-covariance decomposition is a well-defined trade-off \cite{brown2005diversity}.

For simplicity, given the simple average form of the hybrid ensemble $f_h = \frac{1}{m}\sum_{j=1}^{m}f_j$ and the unbiased estimation of the ground truth $\hat{y} = E(y)$, the bias-variance-covariance decomposition is written as the following equation:
\begin{equation}
\label{bias-variance-covariance}
    E((f_h-\hat{y})^2) = B^2+\frac{1}{m}V+(1-\frac{1}{m})C,
\end{equation}
where \emph{B}, \emph{V}, and \emph{C} are the averaged bias, variance, and covariance of each sub-model in the hybrid ensemble. 
The equations for the three terms are as follows:
\begin{equation}
    B = \frac{1}{m}\sum_{j=1}^{m} \left(E(f_j)-\hat{y} \right),
\end{equation}
\begin{equation}
    V = \frac{1}{m}\sum_{j=1}^{m}E\left((f_j-E(f_j))^2\right),
\end{equation}
\begin{equation}
    C = \frac{1}{m(m-1)}\sum_{j=1}^{k}\sum_{k\neq j}E\left[(f_j-E(f_j))(f_k-E(f_k))\right].
\end{equation}

Unlike ambiguity decomposition, the bias-variance-covariance decomposition can reduce the error of the hybrid ensemble by decreasing the covariance without increasing the bias and variance.
Additionally, the covariance term can be negative, implying that negative correlations between sub-models can contribute to the prediction of the hybrid ensemble.

\subsection{Negative Correlation Learning}
\label{Negative Correlation Learning NN ensemble}
Liu has proposed to achieve diversity within an ensemble by NCL \cite{liu1999ensemble}.
They designed NCL as a training method for neural network ensembles. 
It adds a penalty term to the objective function of each network and trains all networks simultaneously and interactively before combining them. 
The purpose of this training pattern is not to obtain multiple accurate and independent neural networks but to capture the correlations and derive sub-networks with negative correlations using penalty terms, which in turn form a robust combination.
Brown also used NCL by adding a heuristic penalty term to the mean squared error as an objective function \cite{brown2005managing}. 
They systematically control the bias-variance-covariance trade-off by optimizing this objective function. 
In addition, they derived a systematic upper bound on the strength of negative correlation, which tended to stabilize as the number of models within the ensemble increased.
As mentioned in Table~\ref{Methods for obtaining diversity at different stages of the ensemble model.} before, there are sub-model training and combination stages in generating an ensemble model. 
The application of NCL in neural network ensembles belongs to the first stage and the objective function for training the sub-model in a typical ensemble is given below:
\begin{equation}
\label{ncl}
 F_j = \zeta_j+ \lambda p_i(n),
\end{equation}

\begin{equation}
\label{p_i(n)}
\begin{aligned}
p_i(n)  & = \frac{1}{n}\sum_{i=1}^{n}(f_j(x_i)-f_h(x_i))\sum_{k\neq j}(f_k(x_i)-f_h(x_i)) \\
& = -\frac{1}{n}\sum_{i=1}^{n}(f_j(x_i)-f_h(x_i))^2.
\end{aligned}
\end{equation}

It is still given that \emph{m} networks in the ensemble and \emph{n} samples in the dataset.
For the $j_{th}$ network, its objective function $F_j$ during training processing is MSE with an NCL penalty term.
In Equation~(\ref{ncl}) and Equation~(\ref{p_i(n)}), $\lambda$ is the negative correlation strength.
When $\lambda$ equals 0, $F_j$ is equivalent to MSE $\zeta_j$, and the higher the $\lambda$, the stronger the negative correlation strength of the objective function.
Previous approaches to increasing model diversity, such as changing the model structure, were mainly implicit.
Contrastingly, the NCL controls model diversity explicitly by adding a penalty term to the objective function using only the parameter $\lambda$. The effect of NCL is to pull the predictions of the sub-models away from the ensemble while drawing the ensemble closer to the actual values \cite{reeve2018diversity}.

\subsection{Interior-point filter line-search algorithm}
The interior-point filter linear-search algorithm has mature applications in many fields as a general-purpose method for solving optimization and programming \cite{simmons2019proactive,carpio2021short, pulsipher2022unifying}.
This optimization algorithm has been well integrated as an Interior Point OPTimizer (IPOPT) solver in \proglang{Gekko} for friendly use, which is designed by Beal \cite{beal2018gekko}.
As an algebraic modeling language, it excels in solving dynamic optimization problems.
Additionally, \proglang{Gekko} is a Python library that integrates model building, analysis tools, and optimization visualization.
Following, we will briefly introduce IPOPT \cite{wachter2006implementation}.

For convenience, researchers are used to writing the objective function and constraints of the optimization problem by adding equation constraints and slack variables in the standard form, as in Equation~(\ref{objective common}): 
\begin{equation}
\label{objective common}
\begin{aligned}
  arg\min _{x \in \mathbb{R}^n} & \ F(x) \\
  s.t. \ c(x) & = 0,  \\
  \qquad  x_i &\geq 0 .
\end{aligned}
\end{equation}

To solve an optimization problem using the interior point method, one adds an auxiliary barrier to Equation~(\ref{objective common}) and, correspondingly, removes the inequality constraint, as in Equation~(\ref{objective barrier}):
\begin{equation}
\label{objective barrier}
\begin{aligned}
  arg\min _{x \in \mathbb{R}^n} \ \phi_\mu(x) & =F(x)-\mu\sum_{i=1}^{m}\ln(x_i) \\
  s.t. \ c(x) & = 0.
\end{aligned}
\end{equation}

As introduced by W{\"a}chter, $\mu$ is a logarithmic barrier term, and $\mu > 0$ \cite{wachter2006implementation}.
As $\mu \rightarrow 0$, the optimization problem~(\ref{objective barrier}) is more likely to converge to an optimal solution.
The solution of Equation~(\ref{objective barrier}) starts with a relatively small $\mu{}$, such as 0.1, and then iterates using the Newton method combined with a linear search.
IPOPT then determines whether the current feasible solution reduces $\phi_\mu(x)$ compared to the previous feasible solution.
In the absence of a feasible solution, IPOPT transforms the problem~(\ref{objective barrier}) into a feasibility restoration phase by finding a feasible solution that minimizes the norm of the constraint violation $\Vert{c(x)}\Vert_1$, temporarily ignoring the objective function, and thus solving it flexibly.
The above steps are repeated, with $\mu{}$ being reduced each time, until the solution of Equation~(\ref{objective barrier}), or the solution satisfying the first-order optimality condition, is found.
All the procedures would be done by \proglang{Gekko} automatically.

\newpage
\section{Hybrid ensemble for regression with negative correlation learning}
\label{Hybrid ensemble for regression with negative correlation learning}
As one of the most fundamental mathematical problems, regression has many well-established models designed from different perspectives.
A hybrid ensemble is a method to solve regression by the weighted average of the predictions of multiple members.
In this study, by introducing NCL in the hybrid ensemble, the sub-models with diversity will be selected, combined, and weighted to improve the prediction accuracy.
Specifically, this section examines these aspects: model pool construction, sub-model training stage, sub-model combination stage, evaluations, and the proposed hybrid ensemble framework.

\subsection{Model pool construction}
\label{Model Pool Construction}
Many ensemble models adopt cross-validation to train homogeneous models and perform majority voting to select models that work well. 
In contrast, this study draws on the conclusion of Mendes-Moreira that heterogeneous models control diversity and perform better than homogeneous candidates in the model training stage \cite{mendes2012ensemble}. 
When constructing the model pool, we chose the models from different methods.
Eleven regression models are selected in this study, including
Simple Linear Regression (SLR) \cite{zou2003correlation}, Ridge Regression (RR) \cite{hoerl1970ridge}, Bayesian Regression (BR) \cite{box2011bayesian}, Stochastic Gradient Descent Regression (SGDR) \cite{jain2018accelerating}, Polynomial Regression (PR) \cite{stigler1974gergonne} from \textbf{Linear methods}; 
Decision Tree Regression (DTR) \cite{wu2008top}, 
Random Forest Regression (RFR) \cite{ho1995random}, and 
Gradient Boosting Decision Tree (GBDT) \cite{friedman2001greedy} from \textbf{Tree-based methods};
Adaptive Boosting Regression (ABR) \cite{solomatine2004adaboost},  Support Vector Regression (SVR) \cite{drucker1997support},  and Multilayer Perceptron Regression (MPR) \cite{rosenblatt1961principles}.
\textbf{Methods}, \textbf{models}, and \textbf{sub-models} will be mentioned several times in this paper, and we have drawn an example in Figure~\ref{The relationship between the regression method, models and model parameters} to distinguish these three terms.

\begin{figure}[htbp]
  \centering
  \includegraphics[scale=.4]{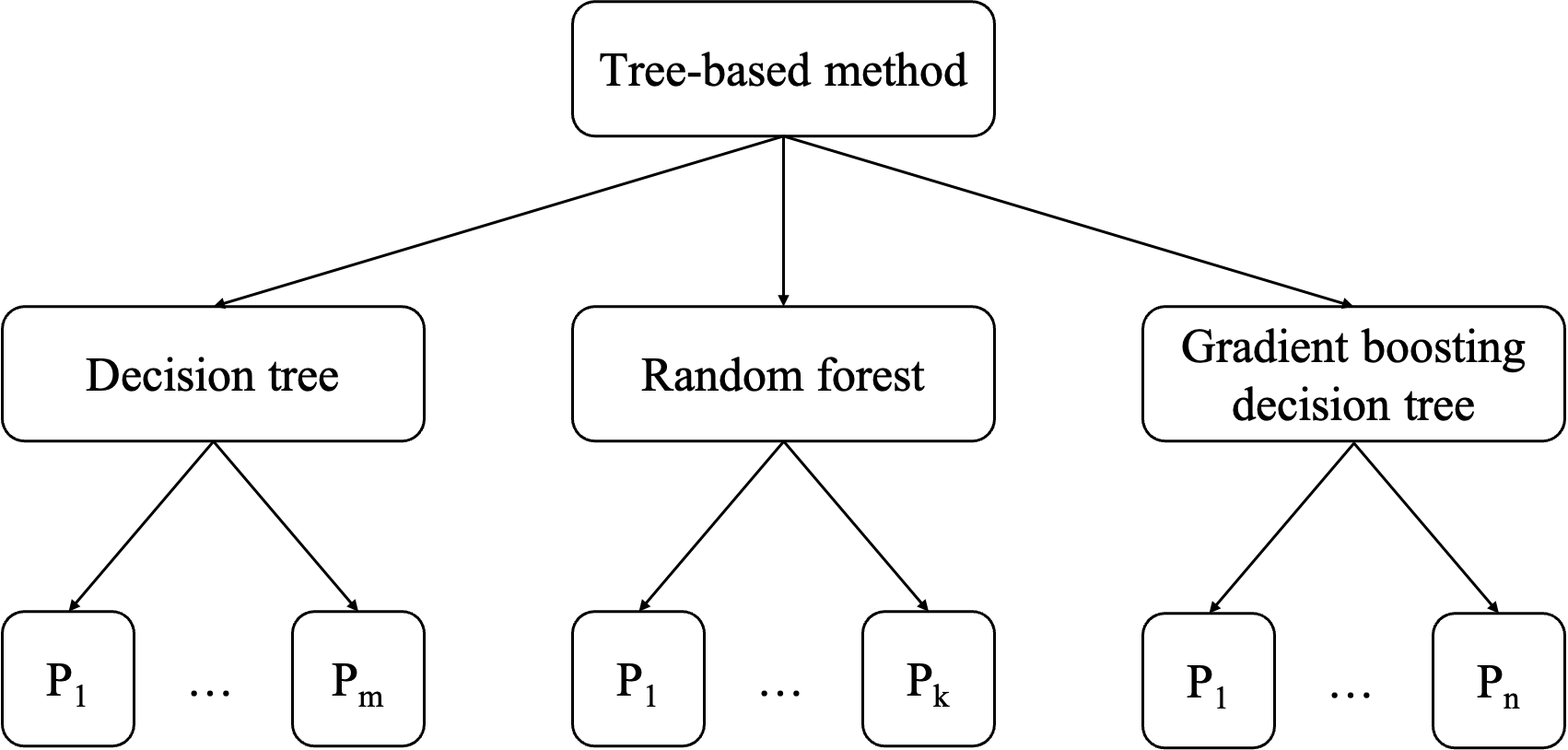}
  \caption{The relationship between the method, models, and sub-models. The top level is a tree-based \textbf{method}, the middle level is different \textbf{models}, and the bottom level are \textbf{sub-models} written as $P_i$ with different parameter sets. The sub-models serve as the members of the hybrid ensembles in this paper.}
  \label{The relationship between the regression method, models and model parameters}
\end{figure}

\subsection{Sub-model training stage}
In practice, grid search is to find the best parameter set of a model to improve the prediction \citep{chicco2017ten}. 
Cross-validation is the basis for judging whether a parameter set is good or not \citep{geisser1975predictive}.
In a typical model fitting task, there will be situations where the training set predicts better than the test set, also known as over-fitting, which can be solved by cross-validation.
In this paper, a 5-fold cross-validation is used, whereby the training data is divided into five equal parts, and a model with a particular parameter set is fitted five times.
The model takes one copy of the data from the training set as the validation set and the remaining four copies as a new training set.
After five fits, the prediction scores on each validation set are averaged as the final score of the current model.
Once the grid search has traversed all possible parameter combinations, the highest-scoring parameter set is taken as optimal.

\begin{figure}[htbp]
  \centering
  \includegraphics[scale=.32]{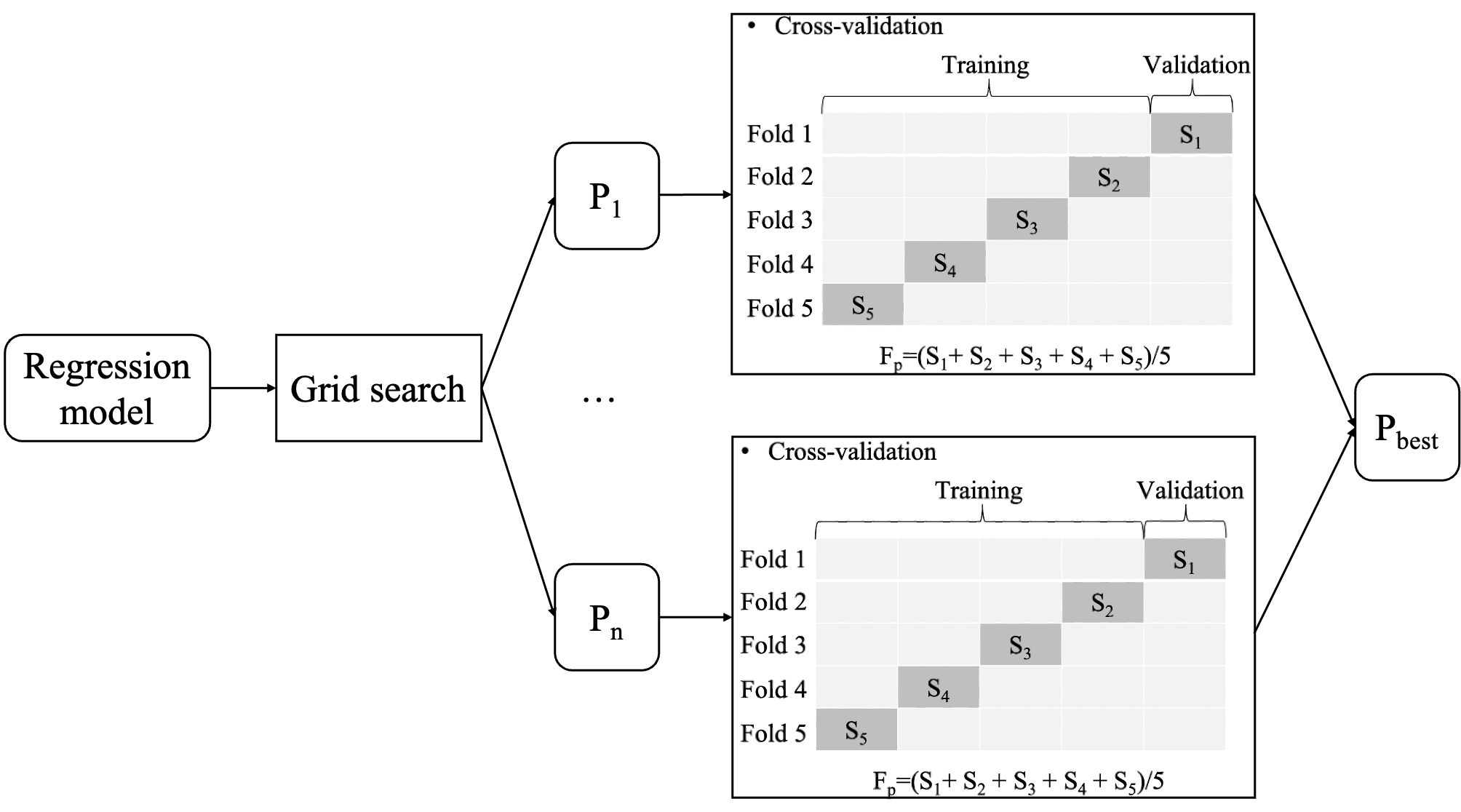}
  \caption{The process of grid search and cross-validation}
  \label{The process of grid search and cross-validation}
\end{figure}

Figure~\ref{The process of grid search and cross-validation} illustrates the process of grid search and cross-validation. 
The value range for each parameter is first set manually to form a discrete parameter space. 
The grid search then traverses the space to obtain all parameter sets, calculates the average prediction error on each validation data, and selects the parameter set with the lowest error.
Once the grid search and cross-validation are finished, we expect to obtain the best parameter set for a model.

\subsection{Sub-model combination stage}
\subsubsection{Objective function for hybrid ensemble}
\label{Objective function for hybrid ensemble}
This subsection explains the difference between the proposed NCL-based hybrid ensemble and the neural network ensemble in \cite{liu1999ensemble} and claims the contributions in detail.
We have introduced how NCL is used in neural network ensemble in Section~\ref{Negative Correlation Learning NN ensemble}.
If combining the Equations~(\ref{ncl}) and (\ref{p_i(n)}), we get the error function for each network:
\begin{equation}
\label{NN ensemble error function}
    F_j = \zeta_j - \frac{\lambda}{n}\sum_{i=1}^n\left( f_j(x_i) - f_h(x_i)\right)^2.
\end{equation}
All network members optimize the error function (\ref{NN ensemble error function}) during training and achieve interaction between members by the penalty term in the function.
The ensemble of networks thus is trained on the `sub-model training' stage in Table~\ref{Methods for obtaining diversity at different stages of the ensemble model.}.

Unlike the neural network ensemble containing homogeneous members, we applied a heterogeneous ensemble to generate the estimators with specialty and accuracy in the different regions of solution space \cite{brown2005diversity}.
Training and interacting sub-models with different architectures in parallel are challenging, so we train each model separately, incorporating diversity in the `sub-model combination' stage.
We consider designing an optimization problem to implement a hybrid ensemble in which candidate sub-models are automatically selected and assigned weights.
We still wrapped the error function of each sub-model as a penalty term to encourage the emergence of diversity as Equation~(\ref{NN ensemble error function}).
Then the objective function of the hybrid ensemble is obtained with the weighted average of all the error functions and is written as follows:
\begin{equation}
\label{objective ncl}
\begin{aligned}
  arg\min _{\omega} \Phi(\omega) & = \sum_{j=1}^{m}\omega_j \Big \{\zeta_j- \frac{\lambda}{n}\sum_{i=1}^n \left(f_j(x_i)-f_h(x_i)\right)^2 \Big \}, \\
  s.t. \ \sum_{j=1}^{m}\omega_j & =1,  \\
   0\leq \omega & \leq 1 .
\end{aligned}
\end{equation}

At this point, we claim the contribution of optimizing Formula~(\ref{objective ncl}) to the performance of the final hybrid ensemble.
Formula~(\ref{objective ncl}) and Formula~(\ref{MSE_HE}), also the ambiguity decomposition,  are similar in form, with the only difference being the $\lambda$ in the second term in Formula~(\ref{objective ncl}).
Further, the Formula~(\ref{MSE_HE}) describes the performance of the hybrid ensemble.
The issue then naturally arises on why hybrid ensemble optimizes Formula~(\ref{objective ncl}) instead of Formula~(\ref{MSE_HE}).
There are three explanations: \textbf{i)} it would be overfitted if one only minimising the Formula~(\ref{MSE_HE}) with focus on the training data;
\textbf{ii)} the ensemble diversity cannot be guaranteed if only the second term of Formula~(\ref{MSE_HE}) is optimized without causing a change in the first term, as both terms contain variance \cite{brown2004diversity}.
\textbf{iii)} Formula~(\ref{MSE_HE}) can be decomposed into the three terms in Formula~(\ref{bias-variance-covariance}) considering the sample distribution. 
The NCL penalty term in Formula~(\ref{objective ncl}) could control the covariance through $\lambda$ without causing bias and variance terms to change and obtain the trade-off between accuracy and diversity.

The differences between the hybrid ensemble and the neural network ensemble can be stated as follows:
\textbf{i)} when training the neural network ensemble, each network has the identical error function as Formula~(\ref{NN ensemble error function}) and interacts with other networks. The network optimization and weight updating are simultaneous.
\textbf{ii)} The proposed ensemble is post-hoc, consisting of heterogeneous sub-models that are pre-tested for the performance on the validation set before being combined. 
This operation avoids the homogeneity of the neural network ensemble but preserves the interaction and enhances the generalization of the hybrid ensemble.
\textbf{iii)} Formula~(\ref{objective ncl}) takes a weighted average of the error functions of all the sub-models instead of optimizing them separately as in the neural network ensemble.
Formula~(\ref{objective ncl}) focuses more on optimizing weights given the known MSE of sub-models on the validation set.
If a sub-model has a higher MSE, Formula~(\ref{objective ncl}) puts less emphasis, or weight, on the sub-model.
The weight can be zero if $\lambda=0$.
However, if this sub-model has a higher difference from the current hybrid ensemble at the same time, it contributes to the diversity of the ensemble and attracts some attention from the Formula~(\ref{objective ncl}).
The penalty term achieves the trade-off between diversity and accuracy with this mechanism.

\subsubsection{Automatic search algorithm for negative correlation penalty}
The $\lambda$ in Formula~(\ref{objective ncl}) controls the strength of the negative correlation penalty.
We designed an algorithm to select a suitable $\lambda$ from a list as Algorithm~(\ref{An automatic search algorithm for optimal}).
The basic idea of searching $\lambda$ is to traverse from 0 to 1 given the step $s$. 
We use \proglang{Gekko} to solve Formula~(\ref{objective ncl}) to obtain weight vector $\omega$ regarding the different sub-models. 
Algorithm~(\ref{An automatic search algorithm for optimal}) then calculates the error of the generated hybrid ensemble on the validation set under $\omega$. 
The error here is a simple average of \emph{RMSE}, \emph{MAE}, and \emph{MAPE}, considering that these three metrics are the evaluation criteria in this paper. 
We attempt to treat these three metrics fairly without preference.
After the algorithm targets the optimal $\lambda^*$ with minimum error, the step $s$ is reduced, and a more refined search is started locally on that $\lambda^*$. 
In this paper, we retain the lambda with three digits, i.e., the search stops when $s<0.001$.

\begin{algorithm}[htbp]
  \caption{An automatic search algorithm for optimal $\lambda^*$}
  \label{An automatic search algorithm for optimal}
  \begin{algorithmic}[1]
   \Input {$\mathcal{F}_v \gets (f_{1},f_{2},...,f_{m})$ \Comment{Predictions on validation set}}
   
   \Initialize{
    $\lambda^* \gets 0.1$ \Comment{Optimal $\lambda$}\\
    $s \gets 0.1$ \Comment{Search step} \\
    $\mathcal{E}^* \gets \infty$ \Comment{Errors on the validation set}
    }
    
    \While {$s>=0.001$} 
        \State $\mathcal{L} \gets [\lambda^*+i*s,\lambda^*-i*s] \backslash \lambda^*$, $i=0,1,..,10$
        \State Remove the $\lambda$ that $\lambda > 1$ or $\lambda < 0$ in $\mathcal{L}$
        \For{$\lambda_i$ in $\mathcal{L}$}
            \State Call \proglang{Gekko} to solve $\Phi(w)$, and obtain the weight $\omega$
            \State Get hybrid ensemble on the validation set: $f_h = \mathcal{F}_t\omega^T$
            \State Compute errors : $\mathcal{E} = (\mathcal{E}_{RMSE}+\mathcal{E}_{MAE}+\mathcal{E}_{MAPE}$)/3
            \If{$\mathcal{E}<\mathcal{E}^*$}
                \State $\mathcal{E}^* \gets \mathcal{E}$
                \State $\lambda^* \gets \lambda_i$
            \EndIf
            \EndFor
        \State $s = s/10$
    \EndWhile
    \Output {Optimal $\lambda^*$ and weights for each sub-model $\omega$}
  \end{algorithmic}
\end{algorithm}

\subsection{Model evaluation metrics}
\label{Model evaluation metrics}
Root Mean Squared Error (\emph{RMSE}), Mean Absolute Error (\emph{MAE}), and Mean Absolute Percentage Error (\emph{MAPE}) are three metrics to evaluate the accuracy of regression models.
The equations of them are as follows with $\hat y_i$ the predicted value, and $y_i$ the true value:
\begin{equation}
\label{fm_rmse}
\emph{RMSE} = \sqrt{\frac{1}{n}\sum_{i=1}^{n}(\hat{y_i}-y_i)^2}
,\end{equation}

\begin{equation}
\label{fm_mae}
\emph{MAE} = \frac{1}{n}\sum_{i=1}^{n}\left\vert \hat{y_i}-y_i \right\vert
,\end{equation}

\begin{equation}
\label{fm_mape}
\emph{MAPE} = \frac{1}{n}\sum_{i=1}^{n}\left\vert \frac{\hat{y_i}-y_i}{y_i} \right\vert
.\end{equation}

\subsection{Framework of the hybrid ensemble}
\label{Framework of the hybrid ensemble}
\begin{figure}[htbp]
  \centering
  \includegraphics[scale=.34]{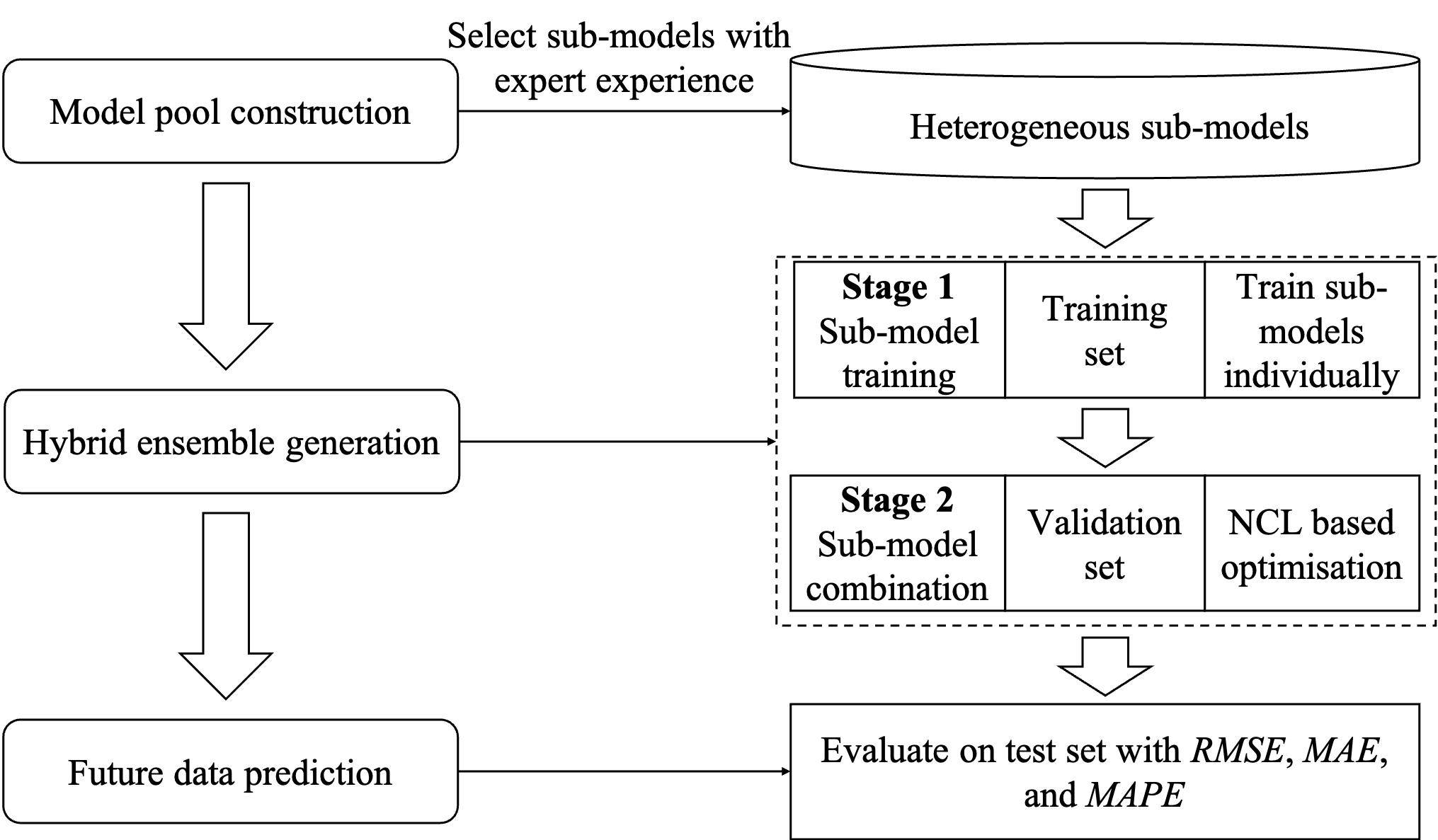}
  \caption{Framework of the hybrid ensemble}
  \label{Framework of the Predictions Combination}
\end{figure}

Figure~\ref{Framework of the Predictions Combination} demonstrates our proposed hybrid ensemble framework incorporating NCL.
This framework includes model pool construction, hybrid ensemble generation, and future data prediction.

Initially, according to expert experience, we select eleven regression sub-models from different aspects like linear models, ensemble models, and neural networks with various structures and parameters to construct the heterogeneous model pool.
Additionally, hybrid ensemble generation contains two stages: sub-model training and sub-model combination.
In the first stage, the training set from the dataset is trained individually by the heterogeneous sub-models in the model pool.
Grid Search and 5-fold Cross-Validation are involved in the training process to find the best parameter set for every model class and avoid overfitting.
In the second stage, the NCL-based objective function is designed for model selection and weighting to find sub-models whose predictions have negative correlations, thus enhancing the diversity within the hybrid ensemble.
The weights of each sub-model are automatically updated in the process of solving the objective function using the IPOPT solver in the \proglang{Gekko} optimizer.
Finally, we treat the test set as future data to evaluate the proposed hybrid ensemble with \emph{RMSE}, \emph{MAE}, and \emph{MAPE}, to see an improvement in contrast to the best-performed sub-model and other benchmarks.

\newpage

\section{Experiments}
\label{Experiments}
This section begins with an introduction of the datasets and model configurations in Section~\ref{Datasets and model configurations}.
Subsequently, we design several sets of experiments from different perspectives to highlight the strength of the proposed approach. 
Section~\ref{Comparison of simple average weighting} starts with the simple average of the elements in each ensemble.
Section~\ref{Construction of NCL ensemble} applies the NCL method on the potential ensembles and explores whether the prediction accuracy will be improved.
Section~\ref{Analysis of sub-model weights} analyses the weights assigned by NCL on sub-models.
Section~\ref{Comparison with state-of-the-art weighting methods} performs the competitive analysis between the NCL ensemble and the state-of-the-art weighting methods.
Section~\ref{Comparison with best sub-model in each group}  compares the prediction effect between the NCL ensembles and the best sub-models in each class.
As the final experimental section, Section~\ref{The sensitivity analysis of negative correlation strength} provides a sensitivity analysis of the negative correlation penalty parameter $\lambda$.

\subsection{Datasets and model configurations}
\label{Datasets and model configurations}
In this study, we chose twenty public datasets from Kaggle\footnote{https://www.kaggle.com/} and UCI machine learning repository\footnote{https://archive.ics.uci.edu/ml/index.php} to test the proposed NCL-based ensemble.
These datasets cover the fields of economy, business, meteorology, and energy.
The names and descriptive statistics are listed in Table~\ref{Description statistics of twenty datasets}.

\begin{table}[htbp]
    \centering
    \caption{Description statistics of twenty datasets}
    \label{Description statistics of twenty datasets}
    \scalebox{0.75}{
        \begin{tabular}{lrrrrrrr}
            \toprule
            Datasets & \# Samples & \# Features & Max of Y & Min of Y & Mean of Y & Median of Y & Std. of Y \\ 
            \midrule
            01-Car & 4,322 & 7 & 8,900,000 & 20,000 & 504,785 & 350,500 & 578,800 \\ 
            02-House & 21,613 & 20 & 7,700,000 & 75,000 & 540,182 & 450,000 & 367,362 \\
            03-Insurance & 1,338 & 6 & 63,770 & 1,121 & 13,270 & 9,382 & 12,110 \\ 
            04-Life\_Expectancy & 2,938 & 21 & 89 & 36 & 69 & 72 & 10 \\
            05-Walmart & 6,435 & 7 & 3,818,686 & 209,986 & 1,046,965 & 960,746 & 564,323 \\ 
            06-Blackfriday & 537,577 & 10 & 23,961 & 185 & 9,334 & 8,062 & 4,981 \\ 
            07-PM25 & 43,824 & 12 & 994 & 0 & 99 & 72 & 92 \\ 
            08-Temperature & 7,752 & 30 & 39 & 17 & 30 & 31 & 3 \\
            09-Power & 9,568 & 4 & 496 & 420 & 454 & 452 & 17 \\ 
            10-Concret & 1,030 & 8 & 82 & 2 & 36 & 34 & 17 \\ 
            11-Gas-2011 & 7,410 & 10 & 119 & 28 & 68 & 66 & 11 \\ 
            11-Gas-2012 & 7,628 & 10 & 120 & 12 & 69 & 67 & 10 \\ 
            11-Gas-2013 & 7,152 & 10 & 120 & 43 & 70 & 69 & 12 \\ 
            11-Gas-2014 & 7,158 & 10 & 118 & 27 & 60 & 59 & 10 \\ 
            11-Gas-2015 & 7,384 & 10 & 120 & 26 & 60 & 57 & 11 \\ 
            12-Traffic & 48,205 & 8 & 7,280 & 0 & 3,260 & 3,380 & 1,987 \\ 
            13-Produce & 1,198 & 14 & 1.12 & 0.23 & 0.74 & 0.77 & 0.17 \\
            14-Election & 21,644 & 27 & 106 & 0 & 1.13 & 0 & 6.87 \\ 
            15-Bike & 8,761 & 13 & 3,556 & 0 & 705 & 505 & 645 \\ 
            16-Steel & 35,041 & 10 & 157 & 0 & 27 & 5 & 33 \\ \bottomrule
        \end{tabular}
    }
\end{table}

Before modeling, data pre-processing is necessary. 
We first removed samples containing null values for each dataset, then transformed nominal variables into one-hot codes and sequential variables into continuous numeric codes.
This paper divided the datasets into training, validation, and test sets. 
In our experiments, the training set was 50\% of the overall. 
When setting the proportion of the validation set, there were two considerations: 
\textbf{i)} the proportion of the validation set cannot be too high. Otherwise, the proportion of the test set would be too low, and the predictions would face a loss of accuracy. 
\textbf{ii)} with a high proportion of validation set, the \proglang{Gekko} solver would not produce a feasible solution due to data overload.
Hence for most of the datasets in this paper, the validation set proportion was 10\% of the total sample. 
As dataset 06-Blackfriday is sufficiently large and dataset 07-PM25 cannot be solved with a validation set ratio of 10\%, the validation set proportions for these two datasets were set to 1\% of the total.

Following this, we set the range of values for the critical parameters of each model. 
The grid search and cross-validation will select the optimal set of parameters from the parameter space for each model. 
The name, parameter range, and the number of sets for each model are listed in Table~\ref{Parameters sets for each model}.
All models and their parameters form the model pool for this paper. 
If the model corresponding to each parameter set is considered a sub-model, the model pool contains 2634 elements.

\begin{table}[htbp]
    \centering
    \caption{Parameters sets for each model}
    \label{Parameters sets for each model}
    \scalebox{0.78}{
        \begin{tabular}{llr}
            \toprule
            Models & Parameters & \# Parameter sets\\ 
            \midrule
            SLR & fit\_intercept:[True,False] & 2 \\ 
            \midrule
            RR	& \makecell[l]{alpha: [0.5,1,2] \\        max\_iter:[100,500,1000] \\ 
                solver:[auto, svd, cholesky, lsqr, sparse\_cg, sag, saga]\\
                tol:[0.0001,0.001,0.01]}
                & 189 \\
            \midrule
            BR	& \makecell[l]{n\_iter:[100,300,500]\\
                tol:[0.0001,0.001,0.01]\\
                alpha\_1:[0.000001,0.0001]\\
                alpha\_2:[0.000001,0.0001]\\
                lambda\_1:[0.000001,0.0001]\\
                lambda\_2:[0.000001,0.0001]\\
                compute\_score:[True,False]\\
                fit\_intercept:[True,False]}	
                & 576\\
            \midrule
            SGDR & \makecell[l]{loss:[squared\_loss,huber,epsilon
            \_insensitive,squared\_epsilon\_insensitive]\\
                penalty:[l1,l2,elasticnet]\\
                alpha:[0.00001,0.0001,0.001]\\
                max\_iter:[500,1000,1500]\\
                tol:[0.0001,0.001,0.01]\\
                learning\_rate:[constant,optimal,invscaling,adaptive]} 
                & 1296 \\
            \midrule
            PR & \makecell[l]{polynomialfeatures\_degree:[2,3]\\
                polynomialfeatures\_interaction\_only:[True,False]\\
                polynomialfeatures\_include\_bias:[True,False]\\
                polynomialfeatures\_order:[C,F]}		
                & 16 \\
            \midrule
            RFR & \makecell[l]{n\_estimators:[50,100,200]\\
                max\_depth:[2,3,4]\\
                min\_samples\_split:[2,3,4]\\
                min\_samples\_leaf:[2,3]\\
                bootstrap:[True,False]}		
                & 108 \\
            \midrule
            ABR	& \makecell[l]{n\_estimators:[10,50,100]\\
                learning\_rate:[0.01,0.1,1]\\
                loss:[linear,square,exponential]}	
                & 27\\
            \midrule
            GBDT & \makecell[l]{n\_estimators:[50,100,200]\\
            learning\_rate:[0.01,0.1,0.5]\\
            loss:[ls,lad,huber,quantile]\\
            min\_samples\_split:[2,3]\\
            max\_depth:[2,3,4]}  
                & 216\\
            \midrule
            SVR & \makecell[l]{kernel:[linear,poly,rbf,sigmoid]\\
                degree:[2,3,4]\\
                C:[0.5,1,2]\\
                gamma:[scale,auto]}		
                & 72\\
            \midrule
            DTR & \makecell[l]{splitter:[best,random]\\
                min\_samples\_split:[2,3]\\
                min\_samples]\_leaf:[2,3]\\
                max\_features:[auto,sqrt,log2]} 
                & 24 \\
            \midrule
            MPR	& \makecell[l]{activation:[identity,logistic,tanh,relu]\\
                solver:[lbfgs,sgd,adam]\\
                alpha:[0.00001,0.0001,0.001]\\
                learning\_rate:[constant,invscaling,adaptive]}
                & 108 \\
            \bottomrule
        \end{tabular}
        }
\end{table}

\subsection{Comparison of simple average weighting}
\label{Comparison of simple average weighting}
Starting with the simple average weighting of sub-models, this section considers the composition of three kinds of ensembles: 
\textbf{i)} an ensemble of all sub-models; 
\textbf{ii)} ensembles of the sub-models within each model class; 
\textbf{iii)} an ensemble of the best sub-models in each model class.

\subsubsection{Diversity of the ensembles}
\label{Diversity of the ensembles}
\begin{figure}[htbp]
  \centering
  \includegraphics[scale=.27]{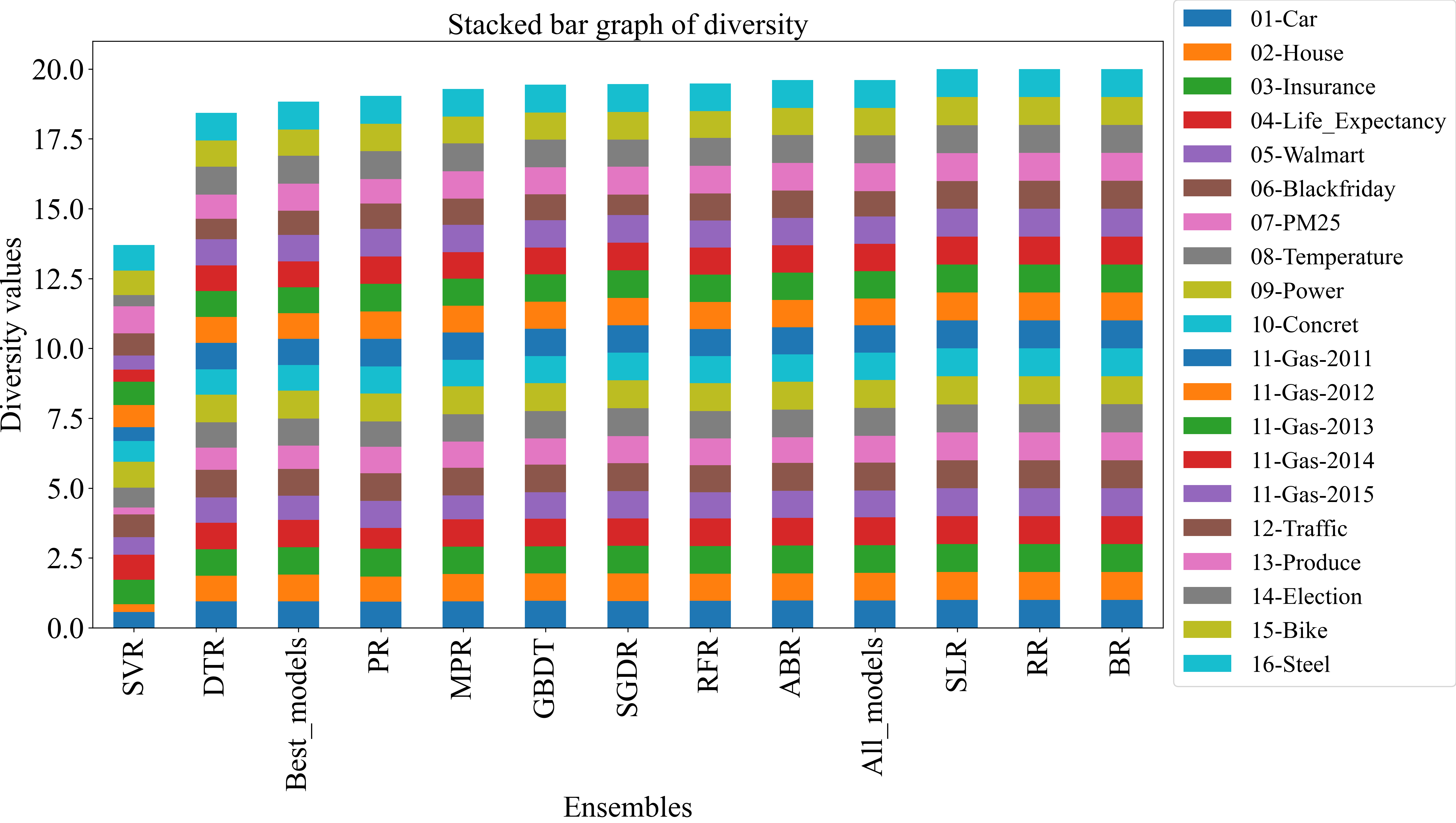}
  \caption{The diversity values across the ensembles in twenty datasets. X-axis is the ensembles. The name \emph{Best\_models} is the ensemble of the best sub-models in each class, \emph{All\_models} is the ensemble of all sub-models, and the rest are ensembles of the sub-models in each class with the same name of the models in Table~\ref{Parameters sets for each model}. Y-axis is the diversity values of each ensemble for all datasets, with the stacked form.}
  \label{the diversity values across the ensembles in twenty datasets}
\end{figure}

Intuitively, the more types of models in an ensemble, the higher the level of diversity. 
In practice, however, it is difficult to define the model types and thus to infer whether the ensemble diversity is caused by the variation of parameters or by the model design itself. 
It has been an opening problem in ensemble learning that needs a consensus diversity measurement.
Nevertheless, we measured the diversity in an ensemble with correlation coefficients as introduced in \cite{dutta2009measuring}.
For several sub-models in an ensemble, we computed the absolute Pearson correlations pairwise and picked the median value as the diversity measurement.
Figure~\ref{the diversity values across the ensembles in twenty datasets} illustrates the diversity values across the ensembles in twenty datasets with stacked bars.

In Figure~\ref{the diversity values across the ensembles in twenty datasets}, the lower values indicate higher diversity since we used the absolute correlations to measure the diversity within an ensemble.
The ensemble \emph{SVR} has the lowest correlation and the highest diversity, followed by \emph{DTR}.
\emph{Best\_models} ranks 3rd and is better than \emph{All\_models} that ranks 10th.
This fact shows that a diverse ensemble does not expect a large amount sub-models.

\subsubsection{Performance of the ensembles}
\label{Performance of the ensembles}
\begin{figure}[htbp]
  \centering
  \includegraphics[scale=.26]{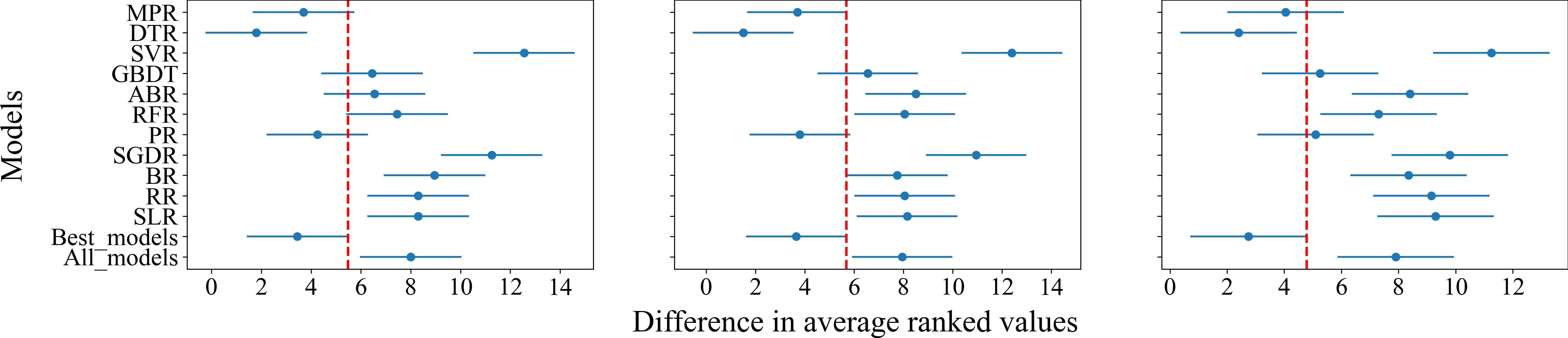}
  \caption{Friedman and Nemenyi test on \emph{RMSE} (left), \emph{MAE} (middle), and \emph{MAPE} (right). The horizontal axis is the differences in average ranked values of each ensemble, and the vertical axis is the names of the ensembles. }
  \label{Friedman_and_Nemenyi_test_Avg}
\end{figure}

To examine the performance of these ensembles statistically, the Friedman and Nemenyi (FN) tests are used in this section \cite{demvsar2006statistical}.
The FN tests are based on the 13 ensembles in Figure~\ref{the diversity values across the ensembles in twenty datasets} ranking on the 20 datasets.
The original hypothesis $\mathcal{H}_0$ of the Friedman test is that all ensembles do not perform significantly differently on all datasets.
If the Friedman test rejects $\mathcal{H}_0$, the Nemenyi test is further used to test whether a significant difference exists between specific ensembles. 
Suppose the difference between the mean ordinal values of the two ensembles is greater than the threshold range of Nemenyi at a certain confidence level. In that case, the predictions of the two ensembles are significantly different.
The results of the FN tests on \emph{RMSE}, \emph{MAE}, and \emph{MAPE} are visualized in Figure~\ref{Friedman_and_Nemenyi_test_Avg}.

In Figure~\ref{Friedman_and_Nemenyi_test_Avg}, each ensemble is represented by a line segment running through a point. 
The points are the average orders of an ensemble over all datasets, and the lower the value on the corresponding horizontal axis, the better the ensemble performs.
The intervals of the line segments are the threshold ranges of the Nemenyi test.
When comparing two ensembles, they are significantly different if there is no overlapping part of their line segments. 
We put red dashed lines in the figures to indicate the maximum average ranked values of the \emph{Best\_models} ensemble.

As can be seen in Figure~\ref{Friedman_and_Nemenyi_test_Avg}, the ensemble \emph{Best\_models} is significantly better than \emph{All\_models}, in line with that the \emph{Best\_models} is more diverse than \emph{All\_models} in Figure~\ref{the diversity values across the ensembles in twenty datasets}.
Another fact is that the ensemble of linear models, except the \emph{PR}, do not offer either high diversity or good performance.
Moreover, we can not tell the significant difference among the ensembles \emph{MPR}, \emph{DTR}, \emph{PR}, and the \emph{Best\_models}.
In Figure~\ref{the diversity values across the ensembles in twenty datasets}, these good-performing ensembles have similar diversity and rank in the top 5.
This phenomenon provides evidence from an experimental perspective that ensemble diversity is associated with performance, whether the ensemble is composed of the same or multiple types of sub-models.
There is still an exception in the ensemble \emph{SVR}, which performs unsatisfactorily compared to the others.
Although it is far more diverse in Figure~\ref{the diversity values across the ensembles in twenty datasets} and there are some cases that \emph{SVR} is the best sub-model in Table~\ref{Comparison with best sub-models}.
This mismatch inspires the future search for the balance between diversity and performance, and an ensemble with excessive diversity may be risky.

\subsection{Construction of NCL ensemble}
\label{Construction of NCL ensemble}
This section constructs an NCL ensemble and compares it with other ensembles designed from different aspects.
In detail, the NCL-based ensemble was built through the best sub-models in each model class, containing eleven members.
The best sub-models were selected by the performance of the model on the validation set.
Then the predictions on the validation set were input into the Algorithm~(\ref{An automatic search algorithm for optimal}) to finish the search for an optimal negative correlation strength $\lambda^*$.
We could also obtain the weights $\omega$ for each sub-model through Algorithm~(\ref{An automatic search algorithm for optimal}).
Finally, we weighted average the predictions on the test set with $\omega$ to generate the final outputs of the NCL-based ensemble.

We would compare the NCL-based ensemble with others considering the sub-model weights~\ref{NCL-based V.S. simple average ensemble}, the ensemble members in Section~\ref{Best sub-models V.S. other ensemble members}, the objective function in Section~\ref{NCL objective function with regularization term}, the training modes in Section~\ref{Hybrid ensemble v.s. neural network}, and the number of sub-models in Section~\ref{The number of sub-models in the NCL ensemble}.

\subsubsection{NCL-based v.s. simple average ensembles}
\label{NCL-based V.S. simple average ensemble}
The difference between NCL-based and simple average ensembles regards the weights of the sub-models.
To explore how the weights influence performance, we compared two ensembles: one with the NCL method paying different attention to each sub-models, the other with equal weights.
Table~\ref{Improvement of the NCL-based ensemble over the simple average} demonstrates the improvement of the NCL-based ensemble over the simple average, in which the metrics with a prefix \emph{Imp} are all measured with percentage.

\begin{table}[htbp]
\centering
\caption{Improvement of the NCL-based ensemble over the simple average}
\label{Improvement of the NCL-based ensemble over the simple average}
\scalebox{0.8}{
\begin{tabular}{lccccc}
\hline
Metrics(\%) & 01-Car        & 02-House    & 03-Insurance   & 04-Life\_Expectancy & 05-Walmart  \\ \hline
ImpRMSE              & 2.21                    & 18.29                & 7.00                       & 9.22                         & 59.86                \\
ImpMAE               & 3.45                    & 19.97                & 14.16                   & 10.72                        & 71.80                 \\
ImpMAPE              & -4.48                   & 12.96                & 15.15                   & 16.25                        & 36.15                \\ \hline
Metrics(\%) & 06-Blackfriday & 07-PM25     & 08-Temperature & 09-Power           & 10-Concret  \\ \hline
ImpRMSE              & 2.57                    & 27.22                & 15.92                   & 1.74                         & 20.11                \\
ImpMAE               & 4.13                    & 31.76                & 17.23                   & 1.87                         & 26.04                \\
ImpMAPE              & -19.42                  & -18.50                & 16.35                   & 9.74                         & 22.35                \\ \hline
Metrics(\%) & 11-Gas-2011    & 11-Gas-2012 & 11-Gas-2013    & 11-Gas-2014        & 11-Gas-2015\\ \hline
ImpRMSE              & 13.68                   & 22.67                & 28.56                   & 20.41                        & 16.44                \\
ImpMAE               & 15.48                   & 23.57                & 31.35                   & 25.93                        & 18.07                \\
ImpMAPE              & 12.13                   & 17.52                & 25.20                    & 48.99                        & 13.24                \\ \hline
Metrics(\%) & 12-Traffic    & 13-Produce & 14-Election  & 15-Bike             & 16-Steel    \\ \hline
ImpRMSE              & 0.17                    & -0.17                & 8.47                    & 20.65                        & 6.82                 \\
ImpMAE               & 0.42                    & 0.17                 & 4.79                    & 26.69                        & 6.85                 \\
ImpMAPE              & -9.45                   & -3.02                & 3.01                    & 6.29                         & 1.56                 \\ \hline
\end{tabular}
}
\end{table}

In Table~\ref{Improvement of the NCL-based ensemble over the simple average}, the NCL-based ensemble could improve the simple average in most cases, around 15\% in \emph{RMSE}, 17\% in \emph{MAE}, and 10\% in \emph{MAPE} on the average of the twenty datasets.
This fact verifies that the ensemble places varying emphasis on its sub-models to enhance performance further, although the sub-models are already diverse.

\subsubsection{Best sub-models v.s. other ensemble members}
\label{Best sub-models V.S. other ensemble members}
Different types of ensemble members are considered here, including \textbf{i)} the best sub-models in each model class, \textbf{ii)} all the sub-models in \emph{DTR}, and \textbf{iii)} the average sub-models in each model class.
The sub-models in \emph{DTR} are chosen for the higher diversity and similar performance as the \emph{Best-models} in Section~\ref{Comparison of simple average weighting}.
The NCL method is used in all three ensembles with different members to generate the weights of sub-models and obtain the final predictions.

Table~\ref{Prediction errors of NCL-based ensembles} presents the prediction errors of the three ensembles.
The \emph{Best-NCL} is the ensemble with the best sub-models of each model class.
The \emph{DTR-NCL} refers to the ensemble with the sub-models in DTR.
The \emph{Mean-NCL} is the ensemble comprising the average predictions generated by each model class.
The values with bold font are the minimum values of error metrics.
                         
\begin{table}[htbp]
    \centering
    \caption{Prediction errors of NCL-based ensembles}
    \label{Prediction errors of NCL-based ensembles}
    \scalebox{0.6}{
        \begin{tabular}{l|ccc|ccc|ccc}
        \hline
        \multirow{2}{*}{Dataset} & \multicolumn{3}{c|}{RMSE}                        & \multicolumn{3}{c|}{MAE}                         & \multicolumn{3}{c}{MAPE} \\
                                 & Best-NCL        & DTR-NCL         & Mean-NCL        & Best-NCL        & DTR-NCL         & Mean-NCL        & Best-NCL        & DTR-NCL         & Mean-NCL \\ 
        \hline
        01-Car                   & 0.698          & \textbf{0.684} & 0.696          & 0.332          & \textbf{0.314} & 0.341          & \textbf{1.989} & 1.990          & 1.999   \\
        02-House                 & \textbf{0.381} & 0.401          & 0.428          & \textbf{0.199} & 0.204          & 0.226          & \textbf{1.313} & 1.403          & 1.456   \\
        03-Insurance             & \textbf{0.348} & 0.388          & 0.377          & \textbf{0.191} & 0.222          & 0.260          & \textbf{0.558} & 0.864          & 0.905   \\
        04-Life\_Expectancy      & 0.265          & \textbf{0.234} & 0.319          & 0.189          & \textbf{0.157} & 0.238          & 1.075          & \textbf{0.802} & 1.551   \\
        05-Walmart               & \textbf{0.254} & 0.284          & 0.321          & \textbf{0.141} & 0.155          & 0.200          & \textbf{1.369} & 1.437          & 1.557   \\
        06-Blackfriday           & 0.718          & 0.712          & \textbf{0.711} & 0.573          & \textbf{0.552} & 0.561          & \textbf{7.415} & 8.131          & 7.960   \\
        07-PM25                  & 0.549          & \textbf{0.500} & 0.526          & 0.360          & \textbf{0.306} & 0.351          & 1.703          & \textbf{1.647} & 1.448   \\
        08-Temperature           & \textbf{0.319} & 0.367          & 0.379          & \textbf{0.240} & 0.273          & 0.290          & \textbf{1.117} & 1.313          & 1.304   \\
        09-Power                 & 0.230          & \textbf{0.208} & 0.247          & 0.178          & \textbf{0.154} & 0.195          & 1.666          & \textbf{1.102} & 1.981   \\
        10-Concret               & \textbf{0.309} & 0.323          & 0.373          & \textbf{0.226} & 0.239          & 0.297          & \textbf{0.798} & 0.839          & 0.833   \\
        11-Gas-2011              & 0.346          & \textbf{0.343} & 0.362          & \textbf{0.198} & 0.207          & 0.226          & \textbf{1.059} & 1.189          & 1.150   \\
        11-Gas-2012              & 0.344          & 0.356          & \textbf{0.338} & 0.224          & 0.238          & \textbf{0.219} & \textbf{1.161} & 1.443          & 1.209   \\
        11-Gas-2013              & \textbf{0.298} & 0.353          & 0.326          & \textbf{0.208} & 0.241          & 0.227          & \textbf{1.071} & 1.313          & 1.149   \\
        11-Gas-2014              & \textbf{0.345} & 0.358          & 0.418          & \textbf{0.211} & 0.222          & 0.268          & \textbf{1.290} & 1.506          & 2.085   \\
        11-Gas-2015              & \textbf{0.290} & 0.294          & 0.328          & 0.200          & \textbf{0.189} & 0.238          & 0.902          & \textbf{0.834} & 0.988   \\
        12-Traffic               & 0.975          & 0.974          & \textbf{0.970} & 0.848          & \textbf{0.843} & 0.844          & \textbf{1.459} & 1.571          & 1.519   \\
        13-Produce               & \textbf{0.498} & 0.537          & 0.551          & \textbf{0.295} & 0.327          & 0.368          & \textbf{0.568} & 0.651          & 0.730   \\
        14-Election              & 0.034          & \textbf{0.029} & 0.044          & 0.011          & \textbf{0.004} & 0.017          & 0.087          & \textbf{0.027} & 0.157   \\
        15-Bike                  & 0.405          & \textbf{0.373} & 0.418          & 0.266          & \textbf{0.236} & 0.284          & 1.079          & \textbf{0.924} & 1.163   \\
        16-Steel                 & 0.075          & \textbf{0.056} & 0.103          & 0.041          & \textbf{0.029} & 0.056          & 0.109          & \textbf{0.100} & 0.149   \\ 
        \hline

        \end{tabular}
    }
\end{table}

\begin{figure}[htbp]
  \centering
  \includegraphics[scale=.75]{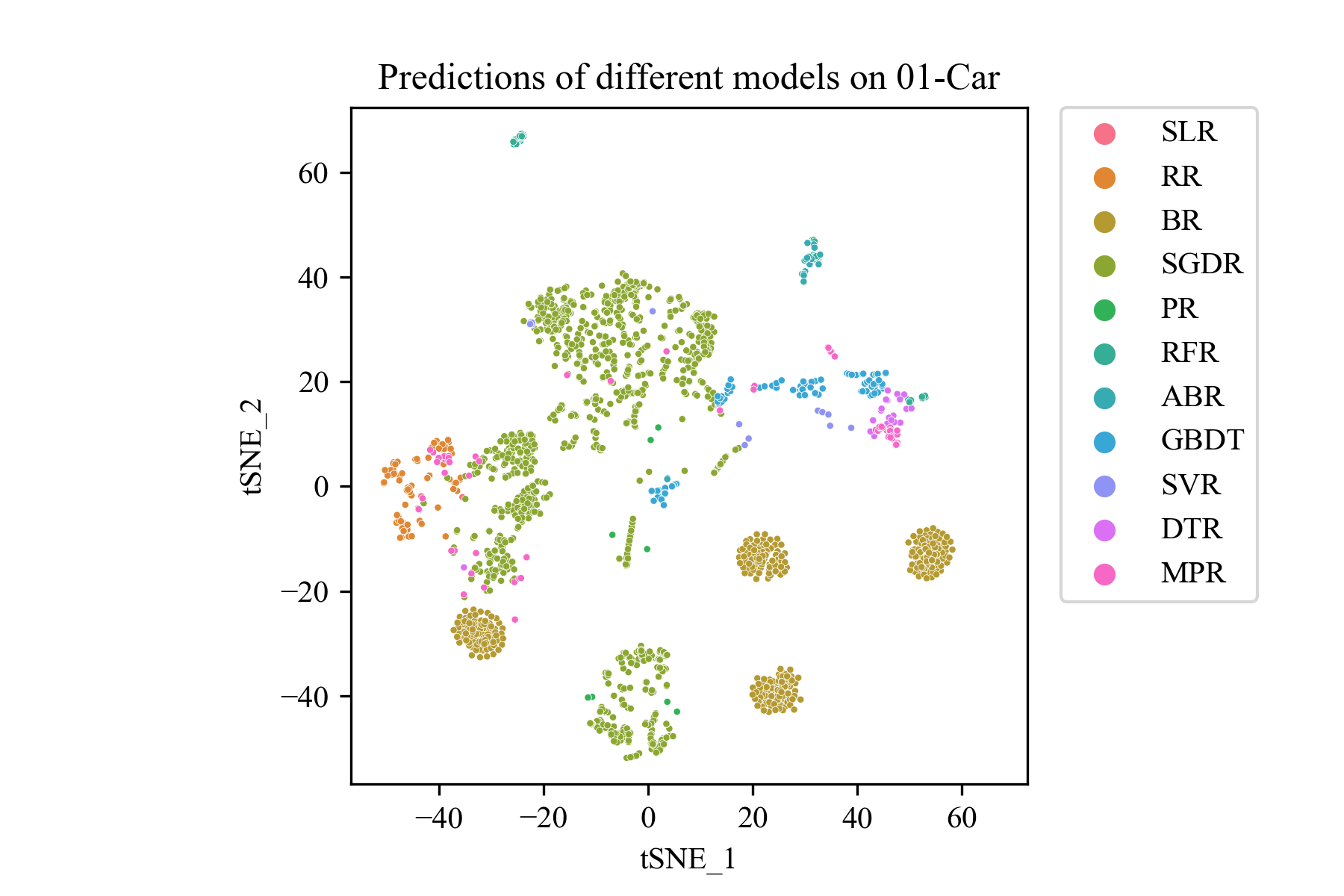}
  \caption{Predictions distribution in the two-dimensional plane of 01-Car}
  \label{Predictions distribution in two-dimensional plane of 01-Car}
\end{figure}

Table~\ref{Prediction errors of NCL-based ensembles} displays that the \emph{Best-NCL} and \emph{DTR-NCL} ensembles take the majority of the minimum errors.
In this case, if an ensemble achieves the minimum error on a dataset, we count it as a win.
The total competition count for each ensemble is 60, with 20 datasets and 3 metrics.
\emph{Best-NCL} wins 31 times out of 60, more than half of them (9, 9, and 13 counts on the three metrics, respectively).
\emph{DTR-NCL} wins 25 times, with 8, 10, and 7 counts on the three metrics.
The \emph{Mean-NCL} wins just 4 times.
The inferior performance of \emph{Mean-NCL} could be explained by the spatial distribution of the predictions from the different models.
Taking `01-Car' as an example, we dimensionalized more than 2000 groups of predictions using the t-SNE technique. 
We visualized them on a two-dimensional plane, as shown in Figure~\ref{Predictions distribution in two-dimensional plane of 01-Car}.

\begin{figure}[htbp]
  \centering
  \includegraphics[scale=.17]{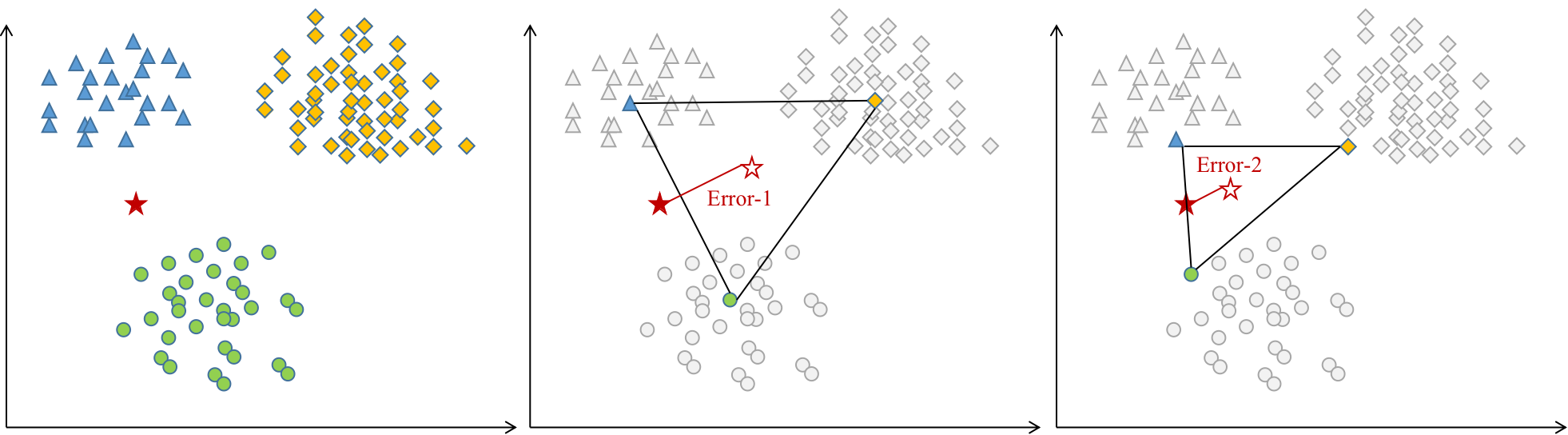}
  \caption{Illustration of the best and average sub-model predictions for each model class}
  \label{Illustration of the best and average sub-model predictions for each model class}
\end{figure}

The visualization provides an intuitive representation of the prediction distributions.
Specifically, predictions generated by one model class with different parameters are distributed in clusters in space and are distinguished from those of other models.
We further abstract this distribution as shown in Figure~\ref{Illustration of the best and average sub-model predictions for each model class}.

In Figure~\ref{Illustration of the best and average sub-model predictions for each model class}, we suppose there are three model classes, each containing several sub-models. 
According to Figure~\ref{Predictions distribution in two-dimensional plane of 01-Car}, the predictions from the same model class form a cluster in the space.
A red star is put in the two-dimensional plane representing the ground truth.
The first sub-plot in Figure~\ref{Illustration of the best and average sub-model predictions for each model class} shows the location of the ground truth and predictions.
The second sub-plot considers the average of the predictions in each class, which locates in the cluster center.
When combining the three cluster centers to form an ensemble, the predictions of the ensemble would be in the center of the triangle region in the sub-plot.
In the third sub-plot, we continue to find the best predictions from each class, which is the point that is nearest to the ground truth.
Then it is evident that the triangle region shrinks as the points are near to the ground truth than the cluster center.
Thus, the ensemble predictions of the best sub-models are closer to the ground truth.

\subsubsection{NCL objective function with regularization term}
\label{NCL objective function with regularization term}
Section~\ref{Objective function for hybrid ensemble} presents the objective function of the NCL ensemble as Formula (\ref{objective ncl}), which includes the MSE and NCL penalty terms.
As pointed out by \cite{chen2009regularized}, the model is easily overfitted when the data has nontrivial noise.
The authors suggest adding a regularization term in the objective function to alleviate the overfitting problem.
Similar as the neural network ensemble in \cite{chen2009regularized}, we redesigned the Formula (\ref{objective ncl}) as follows:

\begin{equation}
\label{objective ncl reg}
\begin{aligned}
  arg\min _{\omega} \Phi(\omega) & = \sum_{j=1}^{m}\omega_j \Big \{\zeta_j- \frac{\lambda}{n}\sum_{i=1}^n \left(f_j(x_i)-f_h(x_i)\right)^2 \Big \} + \sum_{j=1}^m \alpha_j\omega_j^T\omega_j, \\
  s.t. \ \sum_{j=1}^{m}\omega_j & =1,  \\
   0\leq \omega & \leq 1 .
\end{aligned}
\end{equation}

where $\alpha_j$ is the strength of the regularization term $\sum_{j=1}^m\omega_j^T\omega$.
Now we compare the NCL ensemble with and without the regularization term.
The $\alpha_j$ for each sub-models is set equal to 0.05 for simplicity.
Table~\ref{Prediction errors of NCL ensembles with and without regularization term} illustrates the performance of NCL ensembles on the twenty datasets.
The \emph{Best-NCL} is the NCL ensemble without regularization term, and \emph{Best-NCLR} is the NCL ensemble with $\alpha=0.05$.
Table~\ref{Prediction errors of NCL ensembles with and without regularization term} shows that the regularization term marginally improved the NCL ensemble, with more than half of the data sets on each error metric.

\begin{table}[htbp]
\centering
\caption{Prediction errors of NCL ensembles with and without regularization term}
\label{Prediction errors of NCL ensembles with and without regularization term}
\scalebox{0.83}{
\begin{tabular}{l|cc|cc|cc}
\hline
\multirow{2}{*}{Dataset} & \multicolumn{2}{c|}{RMSE} & \multicolumn{2}{c|}{MAE}  & \multicolumn{2}{c}{MAPE}  \\
                         & Best-NCL & Best-NCLR      & Best-NCL & Best-NCLR      & Best-NCL & Best-NCLR      \\ \hline
01-Car                   & 0.698    & 0.780          & 0.332    & 0.397          & 1.989    & 2.358          \\
02-House                 & 0.381    & \textbf{0.368} & 0.199    & \textbf{0.198} & 1.313    & \textbf{1.276} \\
03-Insurance             & 0.348    & 0.359          & 0.191    & 0.194          & 0.558    & 0.560          \\
04-Life\_Expectancy      & 0.265    & \textbf{0.245} & 0.189    & \textbf{0.166} & 1.075    & \textbf{0.941} \\
05-Walmart               & 0.254    & 0.257          & 0.141    & 0.142          & 1.369    & \textbf{1.295} \\
06-Blackfriday           & 0.718    & \textbf{0.714} & 0.573    & \textbf{0.562} & 7.415    & 8.203          \\
07-PM25                  & 0.549    & \textbf{0.544} & 0.360    & \textbf{0.356} & 1.703    & 1.720          \\
08-Temperature           & 0.319    & \textbf{0.307} & 0.240    & \textbf{0.233} & 1.117    & 1.121          \\
09-Power                 & 0.230    & \textbf{0.228} & 0.178    & \textbf{0.176} & 1.666    & \textbf{1.647} \\
10-Concret               & 0.309    & 0.312          & 0.226    & \textbf{0.223} & 0.798    & 0.817          \\
11-Gas-2011              & 0.346    & \textbf{0.345} & 0.198    & 0.198          & 1.059    & \textbf{1.053} \\
11-Gas-2012              & 0.344    & 0.344          & 0.224    & 0.225          & 1.161    & 1.161          \\
11-Gas-2013              & 0.298    & 0.305          & 0.208    & 0.212          & 1.071    & 1.206          \\
11-Gas-2014              & 0.345    & \textbf{0.343} & 0.211    & \textbf{0.209} & 1.290    & \textbf{1.269} \\
11-Gas-2015              & 0.290    & \textbf{0.285} & 0.200    & \textbf{0.193} & 0.902    & 0.927          \\
12-Traffic               & 0.975    & \textbf{0.970} & 0.848    & \textbf{0.845} & 1.459    & \textbf{1.407} \\
13-Produce               & 0.498    & 0.499          & 0.295    & 0.296          & 0.568    & \textbf{0.565} \\
14-Election              & 0.034    & \textbf{0.032} & 0.011    & 0.011          & 0.087    & \textbf{0.086} \\
15-Bike                  & 0.405    & \textbf{0.394} & 0.266    & \textbf{0.255} & 1.079    & \textbf{1.018} \\
16-Steel                 & 0.075    & \textbf{0.070} & 0.041    & \textbf{0.038} & 0.109    & \textbf{0.104} \\ \hline
\end{tabular}}
\end{table}

\subsubsection{Hybrid ensemble v.s. neural network ensemble}
\label{Hybrid ensemble v.s. neural network}
As introduced in Section~\ref{Negative Correlation Learning NN ensemble}, the NCL was developed in the scenario of the neural network ensemble training period.
The hybrid ensemble in this paper transfers the NCL from model training to the combination stage while keeping heterogeneous sub-models as a diverse model pool.
It is still worth comparing the ensembles where NCL works in separate stages.

Given the well-predicted multilayer perceptron in Figure~\ref{Friedman_and_Nemenyi_test_Avg}, we set up a fully connected forward neural network as the sub-model.
After tuning the hyperparameters, we set each sub-network containing two hidden layers, with 16 neurons in each layer. 
The forward propagation took sigmoid as the activation function, and the backward propagation used gradient descent with a regular term to update the weights and biases with a factor of 0.01. 
The individual sub-networks were trained in batches to improve robustness and computational speed, with a batch size of 256.
To match the number of sub-models in the hybrid ensemble, we also set up 11 sub-networks in the network ensemble.
The learning rate of the network ensemble was 0.001, and the negative correlation strength $\lambda$ of both ensembles was 0.5.

Similar to Table~\ref{Improvement of the NCL-based ensemble over the simple average}, Table~\ref{Improvement of the hybrid ensemble over the network ensemble} illustrates the percentage improvement of the hybrid ensemble consisting of the best sub-models over the network ensemble trained by NCL.
In most cases, the hybrid ensemble improves the network ensemble with around 19\% on RMSE, 22\% on MAE, and 25\% on MAPE on the average of all the datasets.
Compared to the simple average, the NCL-based hybrid ensemble achieves a higher percentage improvement over the network ensemble.
The results indicate that even a simple averaged heterogeneous ensemble outperforms a weight-optimized homogeneous ensemble in our regression case.

\begin{table}[htbp]
\caption{Improvement of the hybrid ensemble over the network ensemble}
\label{Improvement of the hybrid ensemble over the network ensemble}
\scalebox{0.85}{
\begin{tabular}{lccccc}
\hline
Metrics(\%) & 01-Car         & 02-House    & 03-Insurance   & 04-Life\_Expectancy & 05-Walmart  \\ \hline
ImpRMSE     & -16.38        & 23.87      & 4.40          & 16.69              & 65.29      \\
ImpMAE      & -22.82        & 27.10      & 13.98         & 21.96              & 73.41      \\
ImpMAPE     & -16.57        & 32.41      & 37.91         & 48.46              & 44.90      \\ \hline
Metrics(\%) & 06-Blackfriday & 07-PM25     & 08-Temperature & 09-Power            & 10-Concret  \\ \hline
ImpRMSE     & 3.07         & 30.77      & 26.19         & 7.07               & 11.62      \\
ImpMAE      & 5.02          & 34.62      & 27.13         & 8.13               & 14.93      \\
ImpMAPE     & -35.31        & 18.43      & 31.38         & 3.82               & 14.15      \\ \hline
Metrics(\%) & 11-Gas-2011    & 11-Gas-2012 & 11-Gas-2013    & 11-Gas-2014         & 11-Gas-2015 \\ \hline
ImpRMSE     & 8.29          & 11.59      & 22.32         & 20.04              & 9.30      \\
ImpMAE      & 9.89          & 13.49      & 20.74         & 23.35              & 7.17       \\
ImpMAPE     & 11.02         & 23.97      & 38.97         & 24.54              & 11.10      \\ \hline
Metrics(\%) & 12-Traffic     & 13-Produce  & 14-Election    & 15-Bike             & 16-Steel    \\ \hline
ImpRMSE     & -0.14         & 8.05       & 71.78         & 23.80              & 42.30      \\
ImpMAE      & -0.55         & 12.89      & 82.31         & 27.35              & 45.23      \\
ImpMAPE     & 6.86          & 22.09      & 81.57         & 46.58              & 60.59      \\ \hline
\end{tabular}}
\end{table}

\subsubsection{The number of sub-models in the NCL ensemble}
\label{The number of sub-models in the NCL ensemble}
In this paper, 11 model classes were initially selected empirically according to the model design and architecture, and the corresponding 11 best sub-models were generated based on the prediction results on the validation set, thereby forming the NCL hybrid ensemble.
Regarding the number of sub-models in the NCL ensemble, we construct the ensemble with the top 5 and top 3 sub-models in each dataset.
Table~\ref{Prediction errors of NCL ensembles with different numbers of sub-models} lists the prediction errors of the ensembles with all the sub-models, the top 5 and top 3 sub-models in each dataset.

\begin{table}[htbp]
\centering
\caption{Prediction errors of NCL ensembles with different numbers of sub-models}
\label{Prediction errors of NCL ensembles with different numbers of sub-models}
\scalebox{0.7}{
\begin{tabular}{l|ccc|ccc|ccc}
\hline
\multirow{2}{*}{Dataset} & \multicolumn{3}{c|}{RMSE}                 & \multicolumn{3}{c|}{MAE}                  & \multicolumn{3}{c}{MAPE}                  \\
                         & NCL-All & NCL-T5         & NCL-T3         & NCL-All & NCL-T5         & NCL-T3         & NCL-All & NCL-T5         & NCL-T3         \\ \hline
01-Car                   & 0.698   & 0.780          & 0.780          & 0.332   & 0.397          & 0.397          & 1.989   & 2.349          & 2.349          \\
02-House                 & 0.381   & 0.388          & 0.388          & 0.199   & 0.209          & 0.209          & 1.313   & 1.336          & 1.336          \\
03-Insurance             & 0.348   & 0.374          & 0.388          & 0.191   & 0.212          & 0.218          & 0.558   & 0.558          & 0.598          \\
04-Life\_Expectancy      & 0.265   & 0.280          & 0.306          & 0.189   & \textbf{0.184} & 0.201          & 1.075   & \textbf{1.001} & \textbf{1.101} \\
05-Walmart               & 0.254   & \textbf{0.248} & 0.599          & 0.141   & 0.143          & 0.424          & 1.369   & \textbf{0.969} & 2.377          \\
06-Blackfriday           & 0.718   & 0.737          & 0.737          & 0.573   & \textbf{0.572} & \textbf{0.572} & 7.415   & 7.873          & 7.873          \\
07-PM25                  & 0.549   & 0.673          & 0.736          & 0.360   & 0.464          & 0.468          & 1.703   & 2.287          & 1.723          \\
08-Temperature           & 0.319   & \textbf{0.307} & 0.332          & 0.240   & \textbf{0.233} & 0.242          & 1.117   & 1.121          & 1.119          \\
09-Power                 & 0.230   & \textbf{0.212} & \textbf{0.205} & 0.178   & \textbf{0.160} & \textbf{0.152} & 1.666   & \textbf{1.290} & \textbf{1.096} \\
10-Concret               & 0.309   & \textbf{0.300} & \textbf{0.300} & 0.226   & \textbf{0.224} & \textbf{0.224} & 0.798   & \textbf{0.785} & \textbf{0.784} \\
11-Gas-2011              & 0.346   & 0.367          & 0.349          & 0.198   & 0.221          & 0.199          & 1.059   & 1.202          & \textbf{0.986} \\
11-Gas-2012              & 0.344   & 0.374          & 0.351          & 0.224   & 0.248          & \textbf{0.215} & 1.161   & 1.557          & 1.290          \\
11-Gas-2013              & 0.298   & 0.321          & 0.321          & 0.208   & 0.224          & 0.225          & 1.071   & 1.314          & 1.315          \\
11-Gas-2014              & 0.345   & \textbf{0.340} & \textbf{0.339} & 0.211   & \textbf{0.206} & \textbf{0.208} & 1.290   & \textbf{1.208} & \textbf{1.251} \\
11-Gas-2015              & 0.290   & \textbf{0.287} & \textbf{0.287} & 0.200   & \textbf{0.196} & \textbf{0.196} & 0.902   & 0.941          & 0.941          \\
12-Traffic               & 0.975   & 0.975          & 0.975          & 0.848   & 0.848          & 0.848          & 1.459   & 1.459          & 1.459          \\
13-Produce               & 0.498   & 0.517          & 0.549          & 0.295   & 0.312          & 0.342          & 0.568   & 0.629          & 0.707          \\
14-Election              & 0.034   & 0.037          & 0.036          & 0.011   & 0.013          & 0.016          & 0.087   & 0.100          & 0.120          \\
15-Bike                  & 0.405   & \textbf{0.384} & 0.410          & 0.266   & \textbf{0.243} & 0.273          & 1.079   & \textbf{0.901} & \textbf{1.078} \\
16-Steel                 & 0.075   & \textbf{0.069} & \textbf{0.055} & 0.041   & \textbf{0.036} & \textbf{0.028} & 0.109   & 0.116          & \textbf{0.099} \\ \hline
\end{tabular}}
\end{table}

In Table~\ref{Prediction errors of NCL ensembles with different numbers of sub-models}, the column name \emph{NCL-ALL} is the NCL ensemble with all the eleven sub-models, the \emph{NCL-T5} and \emph{NCL-T3} correspond to the ensembles with top 5 and top 3 sub-models.
The errors with bolded font are the \emph{NCL-T5} or \emph{NCL-T3} exceeding \emph{NCL-ALL}.
It could be observed that \emph{NCL-T5} or \emph{NCL-T3} performs better than \emph{NCL-All} of each metric only on less than half of the datasets.
This fact leads to the conclusion that the sub-model that constitutes the ensemble is not necessarily the top performer on the dataset. Some of the less-performing sub-models could still contribute negative knowledge to the ensemble, which coincides with the findings of \cite{sirovetnukul2011effectiveness}.

\subsection{Analysis of sub-model weights}
\label{Analysis of sub-model weights}
\begin{figure}[htbp]
  \centering
  \includegraphics[scale=.11]{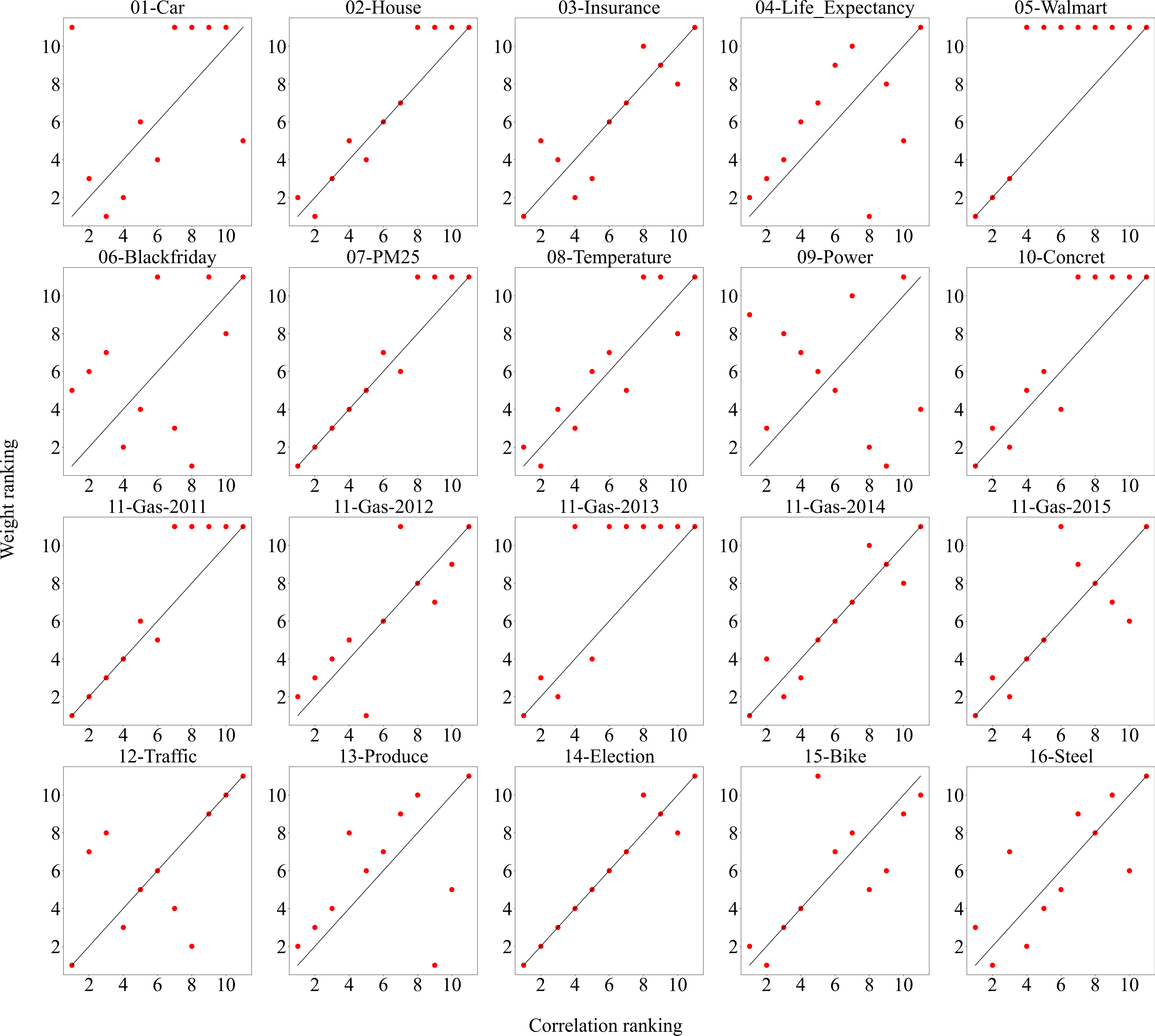}
  \caption{Scatter plot between the rankings of prediction correlation and sub-model weights for 20 datasets. The x-axis is the correlation rankings of the 11 sub-models, and the y-axis is the weight rankings. Each sub-plot is titled by the name of the dataset and contains the scatters as sub-models. The identity line in each subplot indicates that a sub-model has the same ranking in correlation and weight.}
  \label{Scatter plot between the rankings of prediction correlation and sub-model weights}
\end{figure}

As stated in Section~\ref{The number of sub-models in the NCL ensemble}, the components of the ensemble are not necessarily the good-performing sub-models.
In other words, good sub-models may not always contribute to the hybrid ensemble.
In the hybrid ensemble, each sub-model is assigned a weight obtained by the NCL penalty term.
The weights are regarded as the proportion of the sub-models in the ensemble.
This subsection explores whether the weight values of sub-models are related to their ability to contribute to the ensemble.
Concretely, for each dataset, we computed the Pearson correlation coefficients between the predictions of each sub-model and the hybrid ensemble.
The sub-models were ranked according to their correlation with the ensemble from highest to lowest.
Then followed by another ranking list of the sub-model weights from maximum to minimum.
We make scatter plots with the two ranking lists as in Figure~\ref{Scatter plot between the rankings of prediction correlation and sub-model weights}.

Figure~\ref{Scatter plot between the rankings of prediction correlation and sub-model weights} illustrates the relationship between the prediction correlation and weight for each sub-model and dataset.
There are some scatters on the top row of several subplots, such as `01-Car' with five and `02-House' with four.
These scatters correspond to sub-models with zero weights that are filtered out by the NCL ensemble automatically. 
Besides that, the other scatters surround the identity line, exhibiting obvious positive correlations.
These subplots reveal that the higher the weight, the more the sub-model correlates with the ensemble predictions and the more significant its contribution to the final performance.

\subsection{Comparison with state-of-the-art weighting methods}
\label{Comparison with state-of-the-art weighting methods}
The target of a hybrid ensemble is to assign weights to the sub-model with the supervised or unsupervised method.
Besides the NCL-based hybrid ensemble proposed in this study, state-of-the-art methods also assign weights to sub-models.
According to the summary of Mendes-Moreira, there are constant and not-constant weighting methods for building an ensemble \cite{mendes2012ensemble}.
As the name implies, the constant methods assign constant weights to each sub-model. On the other hand, the weights generated by the non-constant methods vary depending on the input data.

\subsubsection{Constant weighting methods}
The most typical constant weighting method is simple averaging, also called the Basic Ensemble Method (BEM) in \cite{mendes2012ensemble}. 
It does not regard the importance of any sub-model nor depend on data attributes and assigns the same weight to all sub-models.
In addition to the simple averaging, the sub-models selected by the NCL-based ensemble are considered here for simple averaging, denoted as BEM-NCL, which has a filtering effect compared to the simple averaging of all sub-models.

Another constant method is Generalized Ensemble Method (GEM) \cite{perrone1992networks}.
GEM generates weights according to the sub-model errors between the actual values and predictions.
In contrast to the BEM, there is no need to assume that these errors are mutually independent and zero-mean.
Let $e_j(x_i) = y_i-f_j(x_i)$ is the error between true value $y_i$ and prediction $f_j(x_i)$ from the $j_{th}$ sub-model.
Then let $w_j$ be the weight assigned on this sub-model, and it is calculated as:
\begin{equation}
    w_j = \frac{\sum_{j=1}^{m}C_{ij}^{-1}}{\sum_{i=1}^{m}\sum_{j=1}^{m}C_{ij}^{-1}},
\end{equation}
in which $C_{ij} = E[e_i(x),e_j(x)]$ is a symmetric correlation matrix of order $M$.

Linear Regression (LR) is also a constant weighting method, with the predictions of the individual sub-models as the independent variables and the true values as the dependent variables.
After the linear regression has fitted the data, the coefficients are taken as the weights for each sub-model.
Unlike GEM, the sum of the linear regression weights does not need to be equal to 1.

\subsubsection{Non-constant weighting methods}
Meta Decision Trees (MDT) method was proposed by \cite{todorovski2003combining} to solve the classification problem, then introduced by \cite{mendes2012ensemble} as a method of non-constant weighting.
MDT is trained on the predictions of the individual sub-models to target true values.
However, it produces a decision tree model rather than a set of coefficients, as in linear regression.
This decision tree model is fitted over the new data to produce the final predicted values, and its potential weights are a decision tree.

Mendes-Moreira classified dynamic weighting, based on the local performance of different sub-models, as a non-constant weighting method \cite{mendes2012ensemble}.
Two intuitive examples are Error Inverse Weighting (EIW) and Error Exponential Weighting (EEW) from Armstrong's design \cite{armstrong2001principles}.
These two weighting methods connect weights to errors, assuming that the higher the error, the less the proportion of the sub-model in the overall ensemble.
The formulas for EIW and EEW are

\begin{equation}
\label{error inverse weights}
    EIW_j = \frac{1/Error_j}{\sum_{j=1}^{m}1/Error_j}, 
\end{equation}

\begin{equation}
\label{error exponential weights}
    EEW_j = \frac{e^{-Error_j}}{\sum_{j=1}^{m}e^{-Error_j}},
\end{equation}
in which $Error_j$ can be any of the metrics from \emph{RMSE}, \emph{MAE}, and \emph{MAPE}.

The NCL-based ensemble proposed in this paper is a dynamic weighting method that integrates model selection with model weighting and belongs to the category of non-constant weighting.

\subsubsection{Comparison with constant and non-constant methods}
After an overview of the classical constant and non-constant weighting methods, this subsection compares the proposed NCL-based ensemble with these weighting methods.
The comparisons between our proposed NCL method (noted as Best-NCL) and the state-of-the-art weighting methods are listed in Table~\ref{Comparison with constant and non-constant methods on RMSE}, Table~\ref{Comparison with constant and non-constant methods on MAE}, and Table~\ref{Comparison with constant and non-constant methods on MAPE}.
These three tables contain the RMSE, MAE, and MAPE results on all twenty datasets.
Besides, we add another row to describe the average ranking of the methods on all datasets in each table.

\begin{table}[htbp]
\centering
    \caption{Comparison with constant and non-constant methods on RMSE}
    \label{Comparison with constant and non-constant methods on RMSE}
    \scalebox{0.85}{
        \begin{tabular}{lcccccccc}
        \hline
Dataset             & BEM            & BEM-NCL        & GEM            & LR             & MDT   & EIW            & EEW            & Best-NCL       \\ \hline
01-Car              & 0.696          & 0.683          & 1.432          & 1.527          & 1.050 & 0.694          & 0.694          & \textbf{0.678} \\
02-House            & 0.463          & 0.423          & 0.454          & 0.466          & 0.521 & 0.446          & 0.454          & \textbf{0.377} \\
03-Insurance        & 0.383          & 0.375          & 0.428          & 0.400          & 0.526 & 0.376          & 0.380          & \textbf{0.355} \\
04-Life\_Expectancy & 0.294          & 0.294          & 0.535          & 0.278          & 0.596 & 0.281          & 0.290          & \textbf{0.266} \\
05-Walmart          & 0.636          & 0.321          & \textbf{0.256} & \textbf{0.256} & 0.404 & 0.492          & 0.563          & \textbf{0.256} \\
06-Blackfriday      & 0.736          & 0.718          & 0.717          & \textbf{0.716} & 0.978 & 0.732          & 0.733          & 0.718          \\
07-PM25             & 0.752          & 0.685          & 0.534          & \textbf{0.533} & 0.758 & 0.725          & 0.732          & 0.549          \\
08-Temperature      & 0.376          & 0.371          & 0.340          & 0.346          & 0.448 & 0.366          & 0.372          & \textbf{0.319} \\
09-Power            & 0.234          & 0.234          & \textbf{0.217} & \textbf{0.217} & 0.278 & 0.231          & 0.233          & 0.230          \\
10-Concret          & 0.380          & 0.329          & 0.396          & 0.363          & 1.001 & 0.357          & 0.369          & \textbf{0.309} \\
11-Gas-2011         & 0.405          & 0.348          & 0.385          & 0.382          & 0.730 & 0.384          & 0.395          & \textbf{0.346} \\
11-Gas-2012         & 0.443          & 0.423          & 0.474          & 0.471          & 0.522 & 0.406          & 0.422          & \textbf{0.344} \\
11-Gas-2013         & 0.419          & 0.319          & 0.369          & 0.395          & 0.484 & 0.381          & 0.400          & \textbf{0.298} \\
11-Gas-2014         & 0.436          & 0.423          & 0.373          & 0.377          & 0.606 & 0.407          & 0.421          & \textbf{0.345} \\
11-Gas-2015         & 0.342          & 0.328          & 0.329          & 0.333          & 0.459 & 0.321          & 0.333          & \textbf{0.290} \\
12-Traffic          & \textbf{0.975} & \textbf{0.975} & 1.030          & 1.047          & 1.324 & \textbf{0.975} & \textbf{0.975} & \textbf{0.975} \\
13-Produce          & \textbf{0.490} & \textbf{0.490} & 0.536          & 0.500          & 0.831 & \textbf{0.490} & \textbf{0.490} & 0.498          \\
14-Election         & 0.036          & 0.036          & \textbf{0.009} & 0.014          & 0.079 & 0.016          & 0.035          & 0.034          \\
15-Bike             & 0.511          & 0.513          & 0.414          & 0.414          & 0.597 & 0.483          & 0.496          & \textbf{0.405} \\
16-Steel            & 0.080          & 0.080          & \textbf{0.031} & \textbf{0.031} & 0.064 & 0.051          & 0.077          & 0.075          \\ \hline
Average ranking     & 5.8            & 3.7            & 3.9            & 3.7            & 7.55  & 3.35           & 4.65           & \textbf{1.85}  \\ \hline
\end{tabular}}
\end{table}

\begin{table}[htbp]
\centering
    \caption{Comparison with constant and non-constant methods on MAE}
    \label{Comparison with constant and non-constant methods on MAE}
    \scalebox{0.85}{
        \begin{tabular}{lcccccccc}
        \hline
Dataset             & BEM            & BEM-NCL        & GEM            & LR             & MDT   & EIW   & EEW   & Best-NCL       \\ \hline
01-Car              & 0.339          & \textbf{0.325} & 0.677          & 0.688          & 0.468 & 0.337 & 0.338 & \textbf{0.325} \\
02-House            & 0.248          & 0.226          & 0.265          & 0.275          & 0.288 & 0.234 & 0.244 & \textbf{0.198} \\
03-Insurance        & 0.229          & 0.225          & 0.266          & 0.199          & 0.298 & 0.213 & 0.225 & \textbf{0.196} \\
04-Life\_Expectancy & 0.213          & 0.213          & 0.383          & 0.197          & 0.448 & 0.198 & 0.209 & \textbf{0.190} \\
05-Walmart          & 0.499          & 0.208          & 0.151          & 0.151          & 0.204 & 0.329 & 0.438 & \textbf{0.142} \\
06-Blackfriday      & 0.597          & 0.574          & \textbf{0.564} & \textbf{0.564} & 0.739 & 0.591 & 0.593 & 0.573          \\
07-PM25             & 0.527          & 0.476          & \textbf{0.347} & \textbf{0.346} & 0.478 & 0.495 & 0.512 & 0.360          \\
08-Temperature      & 0.288          & 0.281          & 0.262          & 0.268          & 0.330 & 0.278 & 0.285 & \textbf{0.240} \\
09-Power            & 0.182          & 0.182          & \textbf{0.160} & \textbf{0.160} & 0.207 & 0.179 & 0.182 & 0.178          \\
10-Concret          & 0.298          & 0.253          & 0.300          & 0.267          & 0.819 & 0.276 & 0.290 & \textbf{0.226} \\
11-Gas-2011         & 0.235          & 0.201          & 0.236          & 0.240          & 0.333 & 0.219 & 0.230 & \textbf{0.198} \\
11-Gas-2012         & 0.293          & 0.277          & 0.302          & 0.302          & 0.355 & 0.267 & 0.284 & \textbf{0.224} \\
11-Gas-2013         & 0.303          & 0.236          & 0.268          & 0.290          & 0.331 & 0.274 & 0.294 & \textbf{0.208} \\
11-Gas-2014         & 0.285          & 0.276          & 0.235          & 0.234          & 0.380 & 0.258 & 0.276 & \textbf{0.211} \\
11-Gas-2015         & 0.242          & 0.235          & 0.234          & 0.236          & 0.332 & 0.224 & 0.236 & \textbf{0.200} \\
12-Traffic          & 0.851          & 0.851          & 0.855          & 0.857          & 1.056 & 0.850 & 0.850 & \textbf{0.848} \\
13-Produce          & \textbf{0.290} & \textbf{0.290} & 0.347          & 0.296          & 0.504 & 0.289 & 0.289 & 0.295          \\
14-Election         & 0.011          & 0.011          & \textbf{0.001} & \textbf{0.001} & 0.014 & 0.002 & 0.011 & 0.011          \\
15-Bike             & 0.363          & 0.354          & 0.287          & 0.287          & 0.443 & 0.332 & 0.351 & \textbf{0.266} \\
16-Steel            & 0.044          & 0.044          & \textbf{0.016} & \textbf{0.016} & 0.028 & 0.024 & 0.043 & 0.041          \\ \hline
Average Ranking     & 5.9            & 3.85           & 4              & 3.6            & 7.35  & 3.45  & 4.7   & \textbf{1.85}  \\ \hline
\end{tabular}}
\end{table}

\begin{table}[htbp]
\centering
    \caption{Comparison with constant and non-constant methods on MAPE}
    \label{Comparison with constant and non-constant methods on MAPE}
    \scalebox{0.85}{
        \begin{tabular}{lcccccccc}
        \hline
Dataset             & BEM   & BEM-NCL        & GEM            & LR             & MDT    & EIW            & EEW            & Best-NCL       \\ \hline
01-Car              & 1.842 & 1.817          & 3.628          & 3.705          & 2.380  & 1.782          & \textbf{1.731} & 1.923          \\
02-House            & 1.449 & 1.476          & 1.757          & 1.818          & 2.050  & 1.347          & 1.285          & \textbf{1.269} \\
03-Insurance        & 0.656 & 0.678          & 0.820          & \textbf{0.515} & 0.651  & 0.580          & 0.597          & 0.555          \\
04-Life\_Expectancy & 1.211 & 1.211          & 2.956          & 1.295          & 6.873  & 1.074          & 1.024          & \textbf{1.013} \\
05-Walmart          & 2.021 & 1.590          & 0.968          & 0.855          & 2.900  & 1.309          & \textbf{0.920} & 1.277          \\
06-Blackfriday      & 6.202 & 7.601          & 7.909          & 7.814          & 14.491 & 4.350          & \textbf{1.966} & 7.415          \\
07-PM25             & 1.434 & 1.649          & 1.926          & 1.916          & 3.112  & 1.360          & \textbf{1.317} & 1.703          \\
08-Temperature      & 1.322 & 1.284          & 1.267          & 1.337          & 2.109  & 1.275          & 1.255          & \textbf{1.117} \\
09-Power            & 1.724 & 1.724          & \textbf{1.303} & 1.311          & 2.452  & 1.583          & 1.499          & 1.666          \\
10-Concret          & 0.999 & 0.820          & 1.040          & 0.954          & 2.409  & 0.927          & 0.913          & \textbf{0.798} \\
11-Gas-2011         & 1.163 & 1.079          & 1.332          & 1.230          & 2.228  & 1.104          & 1.087          & \textbf{1.059} \\
11-Gas-2012         & 1.371 & 1.374          & 2.037          & 2.015          & 2.292  & 1.300          & 1.249          & \textbf{1.161} \\
11-Gas-2013         & 1.452 & \textbf{1.057} & 1.439          & 1.403          & 2.065  & 1.229          & 1.116          & 1.071          \\
11-Gas-2014         & 2.382 & 2.303          & 1.911          & 1.550          & 1.929  & 1.812          & 1.350          & \textbf{1.290} \\
11-Gas-2015         & 1.006 & 1.001          & 1.160          & 1.148          & 1.348  & 0.960          & 0.950          & \textbf{0.902} \\
12-Traffic          & 1.332 & 1.332          & 1.849          & 1.847          & 4.023  & 1.308          & \textbf{1.301} & 1.459          \\
13-Produce          & 0.557 & 0.557          & 0.783          & 0.558          & 1.217  & \textbf{0.545} & 0.551          & 0.568          \\
14-Election         & 0.089 & 0.089          & 0.011          & \textbf{0.010} & 0.019  & 0.021          & 0.085          & 0.087          \\
15-Bike             & 1.355 & 1.346          & 1.112          & 1.110          & 1.360  & 1.196          & 1.125          & \textbf{1.079}          \\
16-Steel            & 0.115 & 0.115          & \textbf{0.077} & \textbf{0.077} & 0.128  & 0.092          & 0.111          & 0.109          \\ \hline
Average Ranking     & 5.2   & 4.55           & 5.4            & 4.5            & 7.4    & 3.25           & \textbf{2.5}   & 2.85           \\ \hline
\end{tabular}}
\end{table}

As illustrated in Table~\ref{Comparison with constant and non-constant methods on RMSE}, Table~\ref{Comparison with constant and non-constant methods on MAE}, and Table~\ref{Comparison with constant and non-constant methods on MAPE}, some remarks can be summarised: i) considering the BEM constant weighting method, NCL is a choice to improve the predictions; ii) Best-NCL performs better than all the methods regarding the number of datasets; iii) the average ranking of Best-NCL is higher than all the methods on RMSE and MAE metrics; iv) on the MAPE metric, the average ranking of Best-NCL is close to that of EEW, although Best-NCL is better than EEW on more datasets. This is caused by the extreme errors on MAPE (06-Blackfriday), while EEW is affected less.

\begin{figure}[htbp]
  \centering
  \includegraphics[scale=.26]{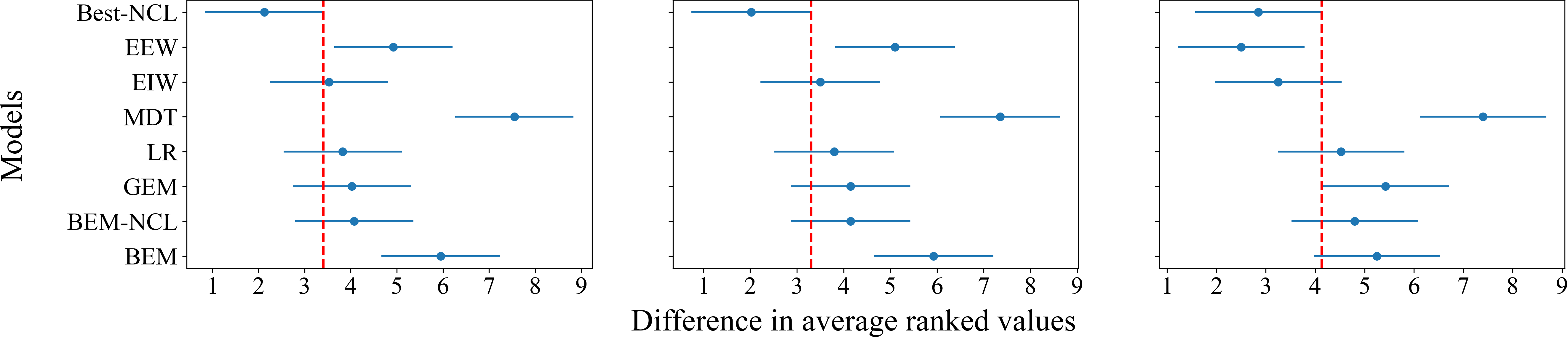}
  \caption{Friedman and Nemenyi test on \emph{RMSE} (left), \emph{MAE} (middle), and \emph{MAPE} (right). The horizontal axis is the differences in average ranked values of each method with the vertical axis the names of them.}
  \label{Friedman_and_Nemenyi_test_Best}
\end{figure}

To significantly show the comparison between the Best-NCL and other methods, the FN tests were performed on the prediction results of the eight weighting methods on the 20 datasets.
Figure~\ref{Friedman_and_Nemenyi_test_Best} presents the results of the statistical analysis of the FN test.

From Figure~\ref{Friedman_and_Nemenyi_test_Best}, our proposed Best-NCL performs better than the other methods on \emph{RMSE} and \emph{MAE} significantly.
In \emph{MAPE}, Best-NCL performs comparably to the two non-constant methods, EIW and EEW, and outperforms the other weighted methods.
These results are in line with what is observed from Table~\ref{Comparison with constant and non-constant methods on RMSE}, Table~\ref{Comparison with constant and non-constant methods on MAE}, and Table~\ref{Comparison with constant and non-constant methods on MAPE}.
All the constant weighting methods listed in this section perform unsatisfactorily, although BEM-NCL with the same sub-models as Best-NCL.
The experiments confirm the superiority of the NCL ensemble and illustrate that fusion of sub-model selection and weighting is necessary when building the ensemble.

\subsection{Comparison with best sub-model in each group}
\label{Comparison with best sub-model in each group}
The previous subsections compared and analyzed NCL-based ensemble with other ensemble methods.
This subsection aims to continue the exploration of the NCL-based ensemble concerning the best sub-models in each model class.
Table~\ref{Comparison with best sub-models} lists the model class that the best sub-model belongs to for each dataset on the validation and test set under the three metrics.
There are columns named `$\lambda=0$' also in Table~\ref{Comparison with best sub-models}, given that the hybrid ensemble only selects one sub-model when there is no NCL.
Bolded fonts in Table~\ref{Comparison with best sub-models} are the sub-models that perform consistently on the validation and test sets. 
If the NCL ensemble outperforms the best sub-model on the final test set, that sub-model is marked with a star.

\begin{table}[htbp]
    \centering
    \caption{Comparison with best sub-models}
    \label{Comparison with best sub-models}
    \scalebox{0.65}{
        \begin{tabular}{l|ccc|ccc|ccc}
        \hline
        \multirow{2}{*}{Dataset} & \multicolumn{3}{c|}{RMSE}                & \multicolumn{3}{c|}{MAE}                 & \multicolumn{3}{c}{MAPE}  
        \\
                                 & Validation set & Test set     & $\lambda=0$ & Validation set & Test set     & $\lambda=0$ & Validation set & Test set     & $\lambda=0$ \\ \hline
        01-Car                   & BR             & DTR          & BR       & RFR            & SVR          & BR       & PR             & BR           & BR       \\
        02-House                 & GBDT           & SVR          & GBDT     & \textbf{DTR}   & \textbf{DTR} & GBDT     & ABR            & SVR          & GBDT     \\
        03-Insurance             & GBDT           & MPR*          & GBDT     & \textbf{MPR}   & \textbf{MPR} & GBDT     & RR             & RFR          & GBDT     \\
        04-Life\_Expectancy      & MPR            & SVR          & DTR      & SVR            & DTR          & DTR      & SVR            & MPR          & DTR      \\
                05-Walmart               & \textbf{MPR}   & \textbf{MPR} & GBDT     & \textbf{MPR}   & \textbf{MPR}* & GBDT     & DTR            & GBDT         & GBDT     \\
        06-Blackfriday           & ABR            & PR*           & PR       & SVR            & ABR          & PR       & RFR            & DTR          & PR       \\
        07-PM25                  & SVR            & GBDT*         & DTR      & \textbf{SVR}   & \textbf{SVR} & DTR      & \textbf{RFR}   & \textbf{RFR} & DTR      \\
        08-Temperature           & MPR            & DTR*          & MPR      & DTR            & SVR*          & MPR      & RFR            & DTR*          & MPR      \\
        09-Power                 & ABR            & DTR          & GBDT     & SVR            & DTR          & GBDT     & PR             & DTR          & GBDT     \\
        10-Concret               & \textbf{MPR}   & \textbf{MPR}* & MPR      & \textbf{MPR}   & \textbf{MPR}* & MPR      & SVR            & MPR*          & MPR      \\
        11-Gas-2011              & \textbf{MPR}   & \textbf{MPR} & SVR      & \textbf{MPR}   & \textbf{MPR} & SVR      & SVR            & GBDT         & SVR      \\
        11-Gas-2012              & \textbf{MPR}   & \textbf{MPR}* & DTR      & DTR            & MPR          & DTR      & GBDT           & MPR          & DTR      \\
                11-Gas-2013              & \textbf{DTR}   & \textbf{DTR}* & MPR      & DTR            & MPR*          & MPR      & SGDR           & MPR          & MPR      \\
        11-Gas-2014              & SVR            & MPR*          & GBDT     & DTR            & MPR*          & GBDT     & RFR            & DTR          & GBDT     \\
        11-Gas-2015              & SVR            & MPR*          & SVR      & DTR            & MPR          & SVR      & RFR            & MPR*          & SVR      \\
        12-Traffic               & PR             & GBDT         & RFR      & SGDR           & ABR          & RFR      & SVR            & DTR          & RFR      \\
        13-Produce               & RFR            & SDGR*         & GBDT     & RFR            & PR           & GBDT     & SVR            & MPR          & GBDT     \\
        14-Election              & \textbf{MPR}   & \textbf{MPR} & DTR      & \textbf{DTR}   & \textbf{DTR} & DTR      & GBDT           & DTR          & DTR      \\
        15-Bike                  & \textbf{DTR}   & \textbf{DTR} & GBDT     & \textbf{DTR}   & \textbf{DTR} & GBDT     & DTR            & SVR          & GBDT     \\
        16-Steel                 & DTR            & MPR          & GBDT     & DTR            & MPR          & GBDT     & GBDT           & DTR          & GBDT     \\ \hline
        \end{tabular}
    }
\end{table}

From Table~\ref{Comparison with best sub-models}, the single sub-model selected by the NCL ensemble when $\lambda=0$ might differ from those in the columns `Validation set' since the objective function takes MSE error.
Table~\ref{Comparison with best sub-models} is an ideal example of model instability.
In the 20 datasets, only a few sub-models perform both best on the validation and test sets.
Model instability occurs when, despite promising results for the current local model, the original optimal model is hard to maintain once new data are available and the data distribution changes.
In some cases, our proposed NCL ensemble is even better than the best-performing sub-models, according to the star marks in Table~\ref{Comparison with best sub-models}, and is a relatively robust ensemble under the RMSE metric.

Building NCL ensembles is a challenging task.
Not only do we have to compare and select sub-models on the validation set thoroughly, but we also have to manipulate the approach to solve the optimization problem, which will undoubtedly consume some time.
However, if the proposed NCL ensemble eventually achieves results comparable to the best-performing sub-model, it means that the ensemble can overcome model instability to some extent. 
This is quite important.
In practice, testing a best-performing sub-model involves picking from a large model pool, which is as time-consuming as building an NCL-based ensemble.
Once the data distribution changes in the future, this sub-model may not continue to predict well, as no perfect single model is suitable for all data.
In this case, the NCL-based ensemble performs more robustly and is a better choice.

\subsection{The sensitivity analysis of negative correlation strength}
\label{The sensitivity analysis of negative correlation strength}
\begin{figure}[htbp]
  \centering
  \includegraphics[scale=.18]{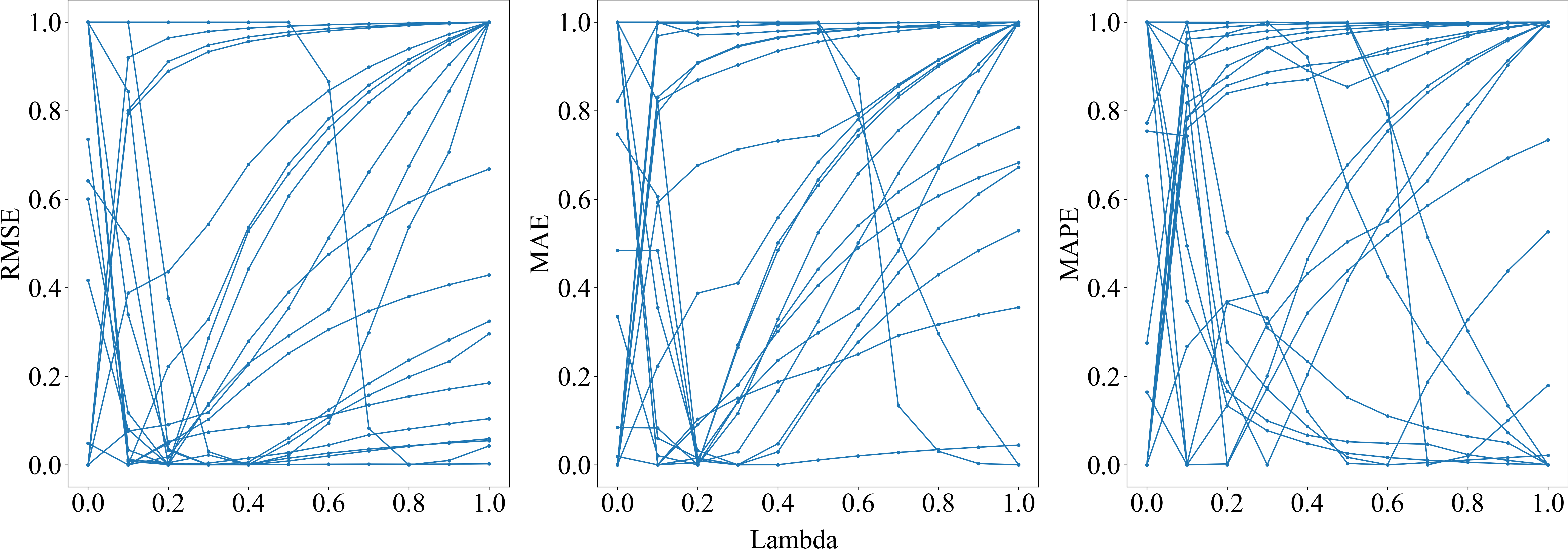}
  \caption{Relationship between ensemble performance and $\lambda$. The horizontal axis is $\lambda$ values from 0 to 1 with step 0.1, and the vertical axis is the corresponding prediction metric \emph{RMSE}, \emph{MAE}, and \emph{MAPE}.}
  \label{Relationship between ensemble performance and lambda}
\end{figure}

The parameter $\lambda$ in the NCL objective function controls the strength of the negative correlation.
If $\lambda$ is close to 0, the NCL objective function can only pick the sub-model with the lowest MSE error on the validation set, which is no different in methodology from selecting a sub-model based on other metrics.
If $\lambda$ is close to 1, The optimization task will search for the most diverse sub-models for the ensemble.
We plot Figure~\ref{Relationship between ensemble performance and lambda} to present how the prediction errors change with $\lambda$.

From Figure~\ref{Relationship between ensemble performance and lambda}, we observe that the \emph{RMSE} and \emph{MAE} errors first decrease at $\lambda$ within 0.1 and 0.2.
This phenomenon shows in more than half of the datasets.
There are also datasets with higher data volume that decreases at higher $\lambda$, such as 06-BlackFriday, and some datasets take lower $\lambda$ and will increase when $\lambda$ is higher than 0.1.
The \emph{MAPE} error is more sensitive than the other two metrics on the change of $\lambda$.
Our findings fit that of Brown \cite{brown2005managing}.
The authors found an upper bound of $\lambda$, and $\lambda$ stabilized when the number of sub-models in the ensemble increased to a certain level.
There is a similar pattern in our sensitivity analysis of $\lambda$.
According to Figure~\ref{Relationship between ensemble performance and lambda}, it is necessary to try different $\lambda$ for different datasets. 
Thus, the Algorithm~\ref{An automatic search algorithm for optimal} designed in this paper to automatically search $\lambda$ is beneficial.

\section{Discussion}
\label{Discussion}
As one of the essential branches of ensemble models, the hybrid ensemble has made significant progress in research and practice.
However, the hybrid ensemble still faces the problem of choosing the appropriate model subset and assigning weights to the sub-models.
Simply averaging the predictions of all sub-models does not achieve the expected results; even the corrections using weighted averaging methods are limited.
This study proposes a novel method for a hybrid ensemble that automatically selects models and generates appropriate weights, yielding comparable performance with the optimal sub-models in regression problems.

A body of studies has experimentally demonstrated that diversity is a critical factor in the success of hybrid ensembles. 
Most studies investigated the sub-model training stage, working on sampling the data and modifying the parameters of homogeneous models, but the diversity generated in this way could be improved. 
This study proposes a regression prediction framework incorporating NCL, considering the diversity in both the sub-model training and combination stages.
Eleven regression models with different structures and parameters are chosen in the sub-model training stage to build a model pool and fit the training set separately. 
Second, the study extends the use of NCL from the previous model training to the model combination. 
Using the interior-point filter linear-search algorithm in the \proglang{Gekko} solver, we solve the optimization problem of model selection and combination to select the negatively correlated model directly sets from the model pool and generate weighted predictions simultaneously. 
Furthermore, the solution to the optimization problem depends on the negative correlation strength $\lambda$. 
Based on this, an algorithm is designed to automatically search for the optimal $\lambda$, avoiding the time wastage of manual search and testing.

The experimental results support that using NCL in the hybrid ensemble is a beneficial initiative, and the importance of diversity is demonstrated in both stages of the ensemble.
In the sub-model training stage, if all the sub-model predictions are projected into a two-dimensional plane, it is evident that those from the same model class will gather into a cluster.
Spatially, the best sub-model in each model class is closer to the true value than the average center of each class. 
The range of geometries formed by the best sub-models is thus more minor, and the ensemble falling in this range has a higher probability of exceeding the average of each model class.
This paper also demonstrates that the ensemble performance is related to the sub-model diversity and that it is statistically better to construct the ensemble with the best sub-models from different model classes.

In the sub-model combination stage, we innovatively considered diversity and solved the optimization problem by incorporating NCL using an interior-point filtering linear-search algorithm.
The experimental results show some inspiring points: \textbf{i)} the NCL ensemble improves the simple average that lacks selecting and weighting procedures; \textbf{ii)} the NCL penalty is beneficial in the sub-models with higher diversity, such as the best sub-models and DTR members; \textbf{iii)} the NCL ensemble can be improved further by adding a regularization term in the objective function; \textbf{iv)} the NCL ensemble performs better than the network ensemble with training the homogeneous sub-models; \textbf{v)} it is necessary to keep some not-satisfying sub-models in the ensemble due to the negative knowledge they may offer; 
\textbf{vi)} the weights of the sub-models are in line with the ensemble performance; 
\textbf{vii)} as a non-constant weighting method, NCL ensemble is superior to other constant weighting methods; 
\textbf{viii)} the NCL ensemble can overcome the model instability and performs close to the best sub-model; 
\textbf{ix)} the auto-searching algorithm is helpful in finding an optimal $\lambda$.

A limitation of this study is that the eleven sub-models in the model pool need to cover more established models in the regression field, which also provides researchers with the freedom to replace candidate models. 
This study also needs a more in-depth exploration of how the data features influence the ensemble effect.

\section{Conclusion}
\label{Conclusion}
We developed a hybrid ensemble approach incorporating negative correlation learning, considering model diversity in the sub-model training and combination stages.
NCL acts as a penalty term for the objective function to be optimized, assisting in the model selection process to find subsets with diversity.
Experiments on twenty publicly available regression datasets confirm the effectiveness of this approach.

First, the proposed method is user-friendly and easy to understand.
Practitioners no longer need to evaluate the effectiveness of individual models using various accuracy indexes to select the best one, nor do they need to blindly weight the candidate models, since the hybrid ensemble with the addition of NCL can fully demonstrate prediction accuracy that approximates or exceeds that of the best sub-model with appropriate penalty strength.
Additionally, the predictions from any model can be added to the model pool as an element for the calculation. 
Even if the model does not work well, this method will discard it automatically. 
Therefore, our proposed method has practical implications.

\section*{Acknowledgement}
We thank the reviewers for their time and efforts in reviewing this manuscript.
We are also grateful for all the valuable comments that help to improve the paper. Yanfei Kang is supported by the National Natural Science Foundation of China (No. 72171011).

\newpage
\section*{Declarations}

\begin{itemize}
\item Funding

Yanfei Kang is supported by the National Natural Science Foundation of China (No. 72171011).

\item Conflict of interest/Competing interests (check journal-specific guidelines for which heading to use)

We declare that we have no financial and personal relationships with other people or organizations that can inappropriately influence our work, and there is no professional or other personal interest of any nature or kind in any product, service and/or company that could be construed as influencing the position presented in, or the review of, the manuscript entitled ``A hybrid ensemble method with negative correlation learning for regression”.

\item Ethics approval 

This paper does not contain any studies with human participants or animals performed by any of the authors.

\item Consent to participate

Informed consent was obtained from all individual participants included in the study.

\item Consent for publication

The author confirms that this publication has been approved by all co-authors.

\item Availability of data and materials

The datasets used in this paper are all from the Kaggle open platform.

\item Code availability

We make our codes publicly on Github\footnote{\url{https://github.com/BaiyunBuaa/Hybrid-ensemble-based-on-Negative-Correlation-Learning}}, please feel free to try!

\item Authors' contributions

Yun Bai: algorithm design, data experiment, and paper writing

Ganglin Tian: data experiment

Yanfei Kang: main theory, paper revision, and submission 

Suling Jia: paper revision and suggestions

\end{itemize}

\clearpage

\appendix

\clearpage
\bibliography{sn-bibliography}

\end{document}